\documentclass[10pt,twocolumn,letterpaper]{article}

\usepackage{iccv}              % To produce the CAMERA-READY version
\usepackage{algorithm}
\usepackage{algpseudocode}

\usepackage{color}
\definecolor{applegreen}{rgb}{0.55, 0.71, 0.0}

\newcommand{\methodname}{\textit{EgoM2P}}

\newcommand{\cmark}{\textcolor{green}{\checkmark}} % Define green checkmark
\newcommand{\halfcmark}{\textcolor{blue}{\checkmark\kern-1.1ex\raisebox{.7ex}{\rotatebox[origin=c]{125}{--}}}} % Define green checkmark
\newcommand{\xmark}{\textcolor{red}{\texttimes}}  % Define red cross
\newcommand{\cmstar}{\textcolor{blue}{\checkmark*}} % Define blue checkmark with star

\definecolor{iccvblue}{rgb}{0.21,0.49,0.74}
\usepackage[pagebackref,breaklinks,colorlinks,allcolors=iccvblue]{hyperref}
\usepackage[accsupp]{axessibility}

\title{EgoM2P: Egocentric Multimodal Multitask Pretraining}

\author{Gen Li$^{1}$ \quad Yutong Chen$^{*1}$ \quad Yiqian Wu$^{*1, 2}$ \quad Kaifeng Zhao$^{*1}$ \quad Marc Pollefeys$^{1, 3}$ \quad Siyu Tang$^{1}$ \vspace{0.3em} \\
{\normalsize $^1$ETH Z\"urich} \quad
{\normalsize $^2$Zhejiang University} \quad
{\normalsize $^3$Microsoft}\\
{\normalsize \url{https://egom2p.github.io/}}
\vspace{-5mm}
}
\begin{document}
\maketitle
\def\thefootnote{*}\footnotetext{Equal contribution. Alphabetical order.}
% \twocolumn[{%
% \renewcommand\twocolumn[1][]{#1}%
% \maketitle
% \begin{center}
%     \centering
%     \captionsetup{type=figure}
%     \includegraphics[width=\textwidth]{img/teaser.pdf}
%     % \fbox{\rule{0pt}{2.5in} \rule{.9\linewidth}{0pt}}
%     \captionof{figure}{\methodname: A large-scale egocentric multimodal and multitask video model, trained on eight extensive egocentric datasets. It incorporates five modalities—RGB and depth video, hand motion, gaze dynamics, and camera trajectories—to handle tasks like monocular egocentric depth estimation, camera tracking, gaze estimation, 4D hand motion reconstruction, and conditional egocentric video synthesis.
% }
%     \label{fig:teaser}
% \end{center}%
% }]

\begin{abstract}
%Egocentric data capture holds great promise for scalable, multimodal human behavior modeling. 
%With the rapid growth of egocentric datasets, there is increasing potential to develop large-scale foundation models for egocentric video, following the success of models trained on third-person-view data. 
Understanding multimodal signals in egocentric vision, such as RGB video, depth, camera poses, and gaze, is essential for applications in augmented reality, robotics, and human-computer interaction, enabling systems to better interpret the camera wearer’s actions, intentions, and surrounding environment. 
However, building large-scale egocentric multimodal and
multitask models presents unique challenges. Egocentric data are inherently heterogeneous, with large variations in modality coverage across devices and settings. 
Generating pseudo-labels for missing modalities, such as gaze or head-mounted camera trajectories, is often infeasible, making standard supervised learning approaches difficult to scale. Furthermore, dynamic camera motion and the complex temporal and spatial structure of first-person video pose additional challenges for the direct application of existing multimodal foundation models.

To address these challenges, we introduce a set of efficient temporal tokenizers and propose \methodname, a masked modeling framework that learns from temporally-aware multimodal tokens to train a large, general-purpose model for egocentric 4D understanding. This unified design supports multitasking across diverse egocentric perception and synthesis tasks, including gaze prediction, egocentric camera tracking, and monocular depth estimation from egocentric video, and also serves as a generative model for conditional egocentric video synthesis. 
Across these tasks, \methodname~matches or outperforms specialist models while being an order of magnitude faster. 
We will fully open-source \methodname~to support the community and advance egocentric vision research.
\end{abstract}    
\vspace{-3mm}
\section{Introduction}
\label{sec:intro}

\begin{figure}[t]
    \centering
    % \setlength{\tabcolsep}{0.0130\linewidth}
    %\begin{tabular}{@{}cc@{}}
    \includegraphics[width=\linewidth]{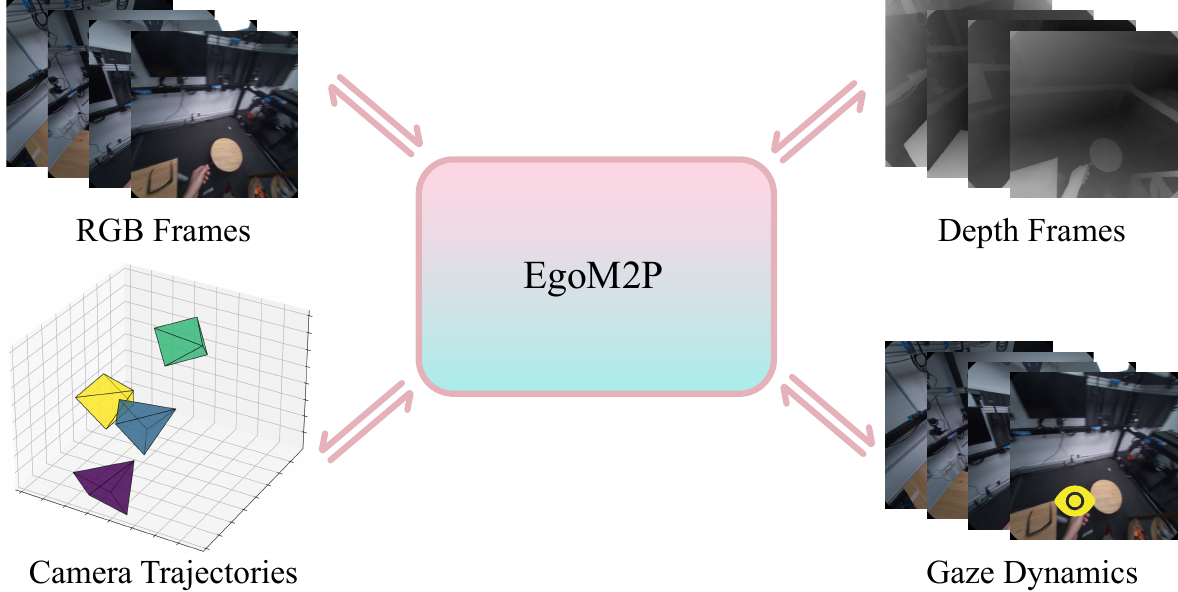}
    % \fbox{\rule{0pt}{2.5in} \rule{.9\linewidth}{0pt}}
    %\end{tabular}
    \caption{\methodname: A large-scale egocentric multimodal and multitask model, pretrained on eight extensive egocentric datasets. It incorporates four modalities—RGB and depth video, 
    % hand motion, 
    gaze dynamics, and camera trajectories—to handle challenging tasks like monocular egocentric depth estimation, camera tracking, gaze estimation, and conditional egocentric video synthesis. For simplicity, we only visualize four frames here. \vspace{-3mm}} 
    \label{fig:teaser} 
\end{figure}

% \siwei{maybe add one sentence motivating why choose these four tasks in introduction, like 'convey most crucial information of ...'}
Egocentric video capture has evolved significantly with the integration of multimodal data, including RGB, depth, gaze, and camera trajectories. 
These modalities interact dynamically, offering the most crucial information for understanding human head motion, intention, and scene geometry.
The growing availability of real-world multimodal egocentric datasets~\cite{Ego4D2022CVPR, Grauman2023EgoExo4DUS, Damen2018EPICKITCHENS, Liu_2022_CVPR, fan2023arctic, Kwon_2021_ICCV, FirstPersonAction_CVPR2018, HoloAssist2023, zhang2022egobody, sener2022assembly101, taco2024, ma2024nymeria, banerjee2024introducing, ohkawa:cvpr23, lv2024aria,avetisyan2024scenescript, 10657079}, such as EgoExo4D, HoloAssist, and HOT3D, provide rich, diverse, and semantically meaningful data.
Additionally, large-scale synthetic datasets generated by simulators like EgoGen \cite{li2024egogen} provide precise ground truth annotations, which are often expensive and time-consuming to obtain in the real world. By combining these complementary data sources, it is increasingly feasible to train large-scale multimodal and multitask models for egocentric vision. These models have the potential to further enhance our understanding of human behavior and real-world interactions, opening new possibilities for applications in augmented reality, virtual reality, and robotics.

Current large-scale video models predominantly focus on understanding and generating videos from third-person perspectives:
Video understanding tasks span video captioning, question answering, retrieval, and segmentation~\cite{wu24next, 10658574, yang2023vid2seq, zhao2023lavila, 10658172, Maaz2023VideoChatGPT, Hu2022RevealRV, yuksekgonul2023when, Zhang2023VideoLLaMAAI};
Video generation models primarily generate third-person view videos from text inputs \cite{runwaygen3, luma, sora, blattmann2023a,guo2023,singer2023, yang2024cogvideox, chen2024videocrafter2,hong2023cogvideo,NEURIPS2022_39235c56,blattmann2023videoldm,wu2023tune,10377444,10377796,villegas2023phenaki,Luo2023VideoFusionDD, wang2023videocomposer}. 
Recent works \cite{moviegen, videopoet} have expanded modalities to include audio, yet these models still operate with a limited range of modalities.
% While these fields advance significantly, egocentric foundation models only inch forward.
While third-person video models have seen significant progress, \textit{egocentric foundation models} have advanced at a much slower pace. 
Egocentric views present unique challenges, including dynamic camera movements, complex human-object interactions, and the influence of human intentions, which are not adequately addressed by third-person models. 
Furthermore, the development of large egocentric models was hindered by the limited scale of available egocentric datasets.
%
% Compared to third-person views, egocentric views are unique due to dynamic camera changes and complex human-scene interactions. Besides, the development is also constrained by the limited egocentric data scale.
With the scaling of egocentric datasets, recent efforts in egocentric foundation models \cite{ye2024mm, suglia2024alanavlm} have focused on egocentric video question-answering (QA). However, these models often overlook critical human-related modalities, which are naturally captured by head-mounted cameras and are essential for understanding human intentions. Additionally, these models lack 3D or 4D reconstruction capabilities, limiting their ability to fully capture the spatiotemporal aspects of human motions and intentions.
% fine-grained human-object interactions.
% and complex human motion patterns.

Recent advances in multimodal and multitask vision foundation models primarily center on images, demonstrating remarkable abilities in cross-modal prediction \cite{4m, 4m21, chen2022a, DBLP:conf/icml/WangYMLBLMZZY22, lu2023unifiedio, Lu2023UnifiedIO2S, kolesnikov2022uvim, beit3}.
These models, which build upon Transformers \cite{NIPS2017_3f5ee243}, enable versatile any-to-any predictions, facilitating multitasking across various modalities, such as depth, surface normal, segmentation masks, etc. However, these models use pseudo-labeling networks to generate aligned binding data across modalities, but effective pseudo-labelers for egocentric videos remain limited due to the domain gap with third-person views.
% \siwei{any reason for that?}. 
Moreover, these models focus on single-image prediction, and struggle to maintain temporal consistency when applied to egocentric video sequences with fast-changing camera poses.

In this paper, we present \methodname, the first multimodal and multitask model for 4D egocentric data. Our approach explores four modalities: RGB video, depth video, gaze dynamics, and camera trajectories. \methodname~supports any-to-any modality predictions and demonstrates its multitasking capabilities across various tasks: gaze estimation, egocentric camera tracking, depth estimation from monocular egocentric video, and conditional egocentric video generation.

Specifically, we build a multimodal token database containing four billion training tokens by curating eight extensive egocentric datasets from both real-world and synthetic data. 
To address missing modality annotations, we effectively extend multimodal masked pretraining, originally designed for image foundation models that assume the availability of all modalities, to the egocentric video domain. While missing modalities are masked out, \methodname~can still effectively predict them. We design a unified temporal tokenizer architecture to tokenize multimodal data into temporally-aware tokens. 
By using variable masking rates to mask input and target tokens during the training of \methodname, we benefit from its parallel inference capability and demonstrate its multitasking performance across various downstream applications, while achieving a significant speed-up.
In summary, the contributions of this work are:
\begin{enumerate}
    \item We introduce \methodname, the first multimodal and multitask large egocentric model for RGB, depth video, eye gaze dynamics, and camera trajectories.
    \item We extend multimodal masked pretraining from the image domain to the egocentric video domain by addressing challenges such as more complex spatiotemporal dynamics and the lack of annotations for certain inherently missing modalities.
    \item \methodname~is comparable to or outperforms state-of-the-art algorithms in egocentric camera tracking, gaze dynamics estimation, monocular egocentric depth estimation, and conditional egocentric video synthesis, while being significantly more efficient. 
\end{enumerate}

\section{Related Work}
\label{sec:formatting}

\noindent\textbf{Image Foundation Models.}
Image foundation models are pretrained versatile neural networks that serve as a universal foundation of various downstream vision tasks, such as image classification, detection, and segmentation~\cite{kirillov2023sam, carion2020detr, liu2022convnet, dosovitskiy2020vit, caron2021dino, oquab2024dinov2}.
% DINO~\cite{caron2021dino, oquab2024dinov2} learns general-purpose visual feature that work across tasks and domains through large-scale self supervised learning.
CLIP~\cite{radford2021CLIP} aligns image and text embeddings via contrastive learning, unlocking zero-shot and open-vocabulary classification.
ImageBind~\cite{girdhar2023imagebind} extends multimodal alignment beyond text, connecting images to modalities like audio, depth, and thermal data.
Multimodal Language Models \cite{2023GPT4VisionSC, team2023gemini, mm1-methods-analysis-insights, abdin2024phi-, liu2024llavanext, wu2024deepseekvl2mixtureofexpertsvisionlanguagemodels, hurst2024gpt} enable unified reasoning across text, images, and audio.
Recent work 4M~\cite{4m, 4m21} trains a multimodal image foundation model using Transformers
% using masked generative transformers~\cite{chang2022maskgit} 
to enable prediction across any input-output modality pairs.
Our work extends 4M by incorporating temporal modeling, training a unified multimodal and multitask model for egocentric vision.
% Repurpose diffusion
% SAPIEN

\noindent\textbf{Video Foundation Models.} 
% Recent advancements in video foundation models build upon the success of Large Language Models (LLMs) \cite{NEURIPS2020_1457c0d6, Achiam2023GPT4TR, touvron2023llama, qwen, dubey2024llama} and Vision Language Models (VLMs) \cite{2023GPT4VisionSC, team2023gemini, mm1-methods-analysis-insights, abdin2024phi-, liu2024llavanext, wu2024deepseekvl2mixtureofexpertsvisionlanguagemodels}, focusing on video understanding and generation through various approaches, such as video-language contrastive learning \cite{xu-etal-2021-videoclip, wang2022internvideo, wang2024internvideo2}, masked modeling \cite{tong2022videomae, wang2023videomaev2, fu2021violet}, and autoregressive sequence prediction \cite{alayrac2022flamingo, sun2024emu, Maaz2023VideoChatGPT, lin2023video, 2023videochat}.
Recent advancements in video foundation models build upon the success of Vision Language Models (VLMs), focusing on video understanding and generation through various approaches, such as video-language contrastive learning \cite{xu-etal-2021-videoclip, wang2022internvideo, wang2024internvideo2},  masked modeling \cite{tong2022videomae, wang2023videomaev2, fu2021violet}, and autoregressive sequence prediction \cite{alayrac2022flamingo, sun2024emu, Maaz2023VideoChatGPT, lin2023video, 2023videochat}.
Diffusion-based video generative models \cite{sora, blattmann2023a, agarwal2025cosmos,  moviegen, kling_video_model} achieve photorealistic video generation with fine-grained content control through conditioning signals such as text prompts and reference images.
These capabilities position video models as promising candidates for world models \cite{agarwal2025cosmos, ha2018world, bruce2024genie, micheli2022transformers, valevski2024diffusion}, as their generative process inherently captures the temporal dynamics of real worlds from internet-scale data.

\noindent\textbf{Egocentric Video Understanding and Generation.}
Understanding the world through egocentric videos is critical for applications in augmented reality, virtual reality, and robotics.
Multiple egocentric video datasets \cite{Ego4D2022CVPR,Grauman2023EgoExo4DUS,Damen2018EPICKITCHENS,HoloAssist2023,banerjee2024introducing} have been collected to capture the diversity and complexity of daily life scenarios.
These egocentric videos present significant technical challenges, including: activity recognition \cite{kazakos2019epic, zhou2015temporal, wang2023ego}, hand motion and object interaction estimation \cite{bambach2015lending,fan2024hold,ye2023diffusion}, egocentric video prediction and generation \cite{liu2024exocentrictoegocentric,liu2021cross, wang2024egovid, xu2025xgen},  egocentric camera localization \cite{teed2021droidslam, sarlin2022lamar}, among others.
While task-specific methods have been developed for these challenges, there remains a lack of a unified egocentric video foundation model.
Our work addresses this gap by enabling cross-modality sequence predictions across RGB, depth, camera poses, and gaze signals, representing a preliminary step toward a foundational multitasking model for a unified egocentric understanding of scenes and human behaviors.
% \noindent\textbf{Spatial AI.}

\section{Method}

\begin{table}[t]
    \centering
    \setlength{\tabcolsep}{2.5pt} % Reduce column spacing
    \renewcommand{\arraystretch}{1.} % Adjust row spacing
    \begin{tabular}{l|cccc}
        \toprule
        \diagbox[width=8em, height=2em]{\scriptsize\textbf{Datasets}}{\scriptsize\textbf{Modalities}} 
        & RGB & Depth & Gaze & Camera \\
        \midrule
        EgoExo4D \cite{Grauman2023EgoExo4DUS} & \cmark & \xmark   & \cmark & \cmark \\
        HoloAssist \cite{HoloAssist2023}    & \cmark & \cmstar & \cmark & \cmark \\
        HOT3D (Aria) \cite{banerjee2024introducing} & \cmark & \cmstar  & \cmark & \cmark \\
        HOT3D (Quest) \cite{banerjee2024introducing} & gray & \cmstar  & \xmark & \cmark \\
        ARCTIC \cite{fan2023arctic}         & \cmark & \cmstar   & \xmark & \cmark \\
        TACO \cite{taco2024}                & \cmark & \cmstar  & \xmark & \cmark \\
        H2O \cite{Kwon_2021_ICCV}           & \cmark & \cmark    & \xmark & \cmark \\
        % Re:InterHand \cite{moon2023reinterhand} & \cmark & \xmark   & \cmark & \xmark & \xmark \\
        % OakInk2 \cite{10657079} & \cmark & \xmark   & \cmark & \xmark & \cmark \\
        % AssemblyHands \cite{ohkawa:cvpr23} & gray & \xmark   & \cmark & \xmark & \cmark \\
        % AEA \cite{lv2024aria} & \cmark & \xmark   & \xmark & \cmark & \cmark \\
        % ASE \cite{avetisyan2024scenescript} & \cmark & \cmark   & \xmark & \xmark & \cmark \\
        EgoGen \cite{li2024egogen} & \cmark & \cmark  & \xmark & \cmark \\
        \bottomrule
    \end{tabular}
    \caption{\textbf{Datasets used in our method.} A green checkmark (\cmark) signifies availability, a red cross (\xmark) denotes unavailability or exclusion due to low quality, and a blue checkmark with a star (\cmstar) indicates the use of pseudo labels. \vspace{-3mm}}
    \label{tab:datasets}
\end{table}

% We detail the data curation pipeline and training paradigm in this section. In Sec.~\ref{sec:curation}, we curate heterogeneous egocentric datasets into unified formats. In Sec.~\ref{sec:tokenizer}, we compress high-dimensional multimodal data into compact discrete tokens for efficient training and inference. In Sec.~\ref{sec:mask_pretrain}, we embed multimodal discrete tokens and perform masked pretraining of \methodname. In Sec.~\ref{sec:generation}, we predict final target tokens by sampling from the pretrained \methodname. \siwei{each sentence starts with 'we did sth...', try to vary the phrase}

This section overviews the data curation pipeline and training paradigm. In Sec.~\ref{sec:curation}, heterogeneous egocentric datasets are transformed into unified formats. Sec.~\ref{sec:tokenizer} describes the process of compressing high-dimensional multimodal data into compact discrete tokens, enabling efficient training and inference. Next, Sec. \ref{sec:mask_pretrain} covers the embedding of multimodal discrete tokens and the masked pretraining of \methodname. Finally, Sec.~\ref{sec:generation} explains how final target tokens are predicted by sampling from the pretrained model.

\subsection{Data Curation}
\label{sec:curation}

As shown in Tab.~\ref{tab:datasets}, egocentric datasets 
differ in their data annotation coverage. Due to hardware and processing constraints, frame drops are common in captured depth data, making it difficult for wearable AR glasses \cite{HoloLens, aria} to achieve pixel-aligned depth streams. Besides, helmet-mounted Kinect cameras are unable to capture gaze information. This unstructured nature of egocentric data poses challenges for both data curation and model training.

Our data curation pipeline includes 3 steps: 1) splitting, 2) annotation, and 3) standardization:
% In total, we process about \gen{todo} hours of egocentric multimodal data.

\noindent\textbf{Splitting.} 
The raw multimodal data are segmented into clips of $T$ frames.
% ; data shorter than 1.5 seconds are discarded, while data longer than 1.5 seconds are padded to $T$ frames. 
These video clips are re-encoded into high-quality mp4 format with the same resolutions.

\noindent\textbf{Annotation.} Real-world egocentric depth data is scarce, often contains sensor noise, and suffers from frame drops. 
We leverage RollingDepth \cite{ke2024rollingdepth} to generate pseudo-labels for the depth annotation to get pixel-aligned depth videos. 
To further scale up the amount of accurate depth training data, we use EgoGen \cite{li2024egogen}, a novel egocentric synthetic data generator, to generate approximately 30 hours of video data at 30 FPS.
This involves letting virtual humans walk in Replica \cite{replica19arxiv} scenes and GIMO \cite{zheng2022gimo} scene scans, rendering their egocentric views to obtain accurate depth and camera trajectory annotations. 
As analyzed in Supp. Mat. Sec. B.3, EgoGen boosts \methodname~performance.
For datasets without gaze annotations, we leave them unlabeled due to the lack of effective pseudo-labelers for gaze dynamics.

\noindent\textbf{Standardization.} These datasets have various video resolutions and define their world coordinate systems differently, with varying origins, axis conventions, handedness, and scale.
We standardize multi‐modal data as follows. All data streams are at 30 FPS. Depth videos are encoded using inverse depth representation, with normalization applied per sequence. Noisy Kinect depth labels are preprocessed with hole filling via morphological operations and inpainting, then denoised using median and bilateral filtering.
% Hand motions are canonicalized in a local coordinate system centered at the first frame. Given the direction from the wrist to the metacarpophalangeal (MCP) joint of the middle finger, as well as the direction of gravity, the canonicalized origin is located at the wrist joint, and the local axis system can be derived by Alg.~\ref{alg:canonical}.
We reproject the eye gaze data, originally represented as a 3D ray with depth, onto the 2D image plane and represent it as a moving point on this plane.
Camera trajectories are unified as camera-to-world transformations in OpenCV convention, using the first frame as the reference frame to standardize the world coordinates across different datasets.

\iffalse
\begin{algorithm}[t]
\caption{Hand Motion Canonicalization}
\begin{algorithmic}
    \State \textbf{Input:} $\mathbf{p}_{\text{middle-MCP}} \in \mathbb{R}^3$, $\mathbf{p}_{\text{wrist}} \in \mathbb{R}^3$
    \State $\mathbf{y} \gets \mathbf{p}_{\text{middle-MCP}} - \mathbf{p}_{\text{wrist}}$
    \State $\mathbf{y} \gets [y_1, y_2, 0]$ \Comment{Project to the xy plane}
    \State $\mathbf{y}_{\text{axis}} \gets \frac{\mathbf{y}}{\|\mathbf{y}\|}$ 
    \State $\mathbf{z}_{\text{axis}} \gets [0, 0, 1]$ \Comment{Inverse gravity direction}
    \State $\mathbf{x} \gets \mathbf{y}_{\text{axis}} \times \mathbf{z}_{\text{axis}}$
    \State $\mathbf{x}_{\text{axis}} \gets \frac{\mathbf{x}}{\|\mathbf{x}\|}$ 
    \State \textbf{Return:} $[\mathbf{x}_{\text{axis}}, \mathbf{y}_{\text{axis}}, \mathbf{z}_{\text{axis}}]$
\end{algorithmic}
\label{alg:canonical}
\end{algorithm}
\fi

\subsection{Tokenizers}
\label{sec:tokenizer}

Given a multi-modal clip with \( T \) frames, we represent each modality as follows:
\begin{itemize}
    \item \textbf{RGB Video}: \( \mathbf{X}^{\text{rgb}} \in \mathbb{R}^{T \times H \times W \times 3} \), where \( H \) and \( W \) denote the spatial resolution.
    \item \textbf{Depth Video}: \( \mathbf{X}^{\text{depth}} \in \mathbb{R}^{T \times H \times W \times 1} \).
    % \item \textbf{Hand Motion}: \( \mathbf{X}^{\text{hand}} \in \mathbb{R}^{T \times J \times 3} \), where \( J \) is the number of hand joints, and each joint is represented in the canonicalized 3D space.
    \item \textbf{Gaze Dynamics}: \( \mathbf{X}^{\text{gaze}} \in \mathbb{R}^{T \times 2} \), where each entry corresponds to the 2D gaze coordinates.
    \item \textbf{Camera Trajectory}: \( \mathbf{X}^{\text{cam}} \in \mathbb{R}^{T \times 9} \), where each pose is parameterized by the 6D rotation representation \cite{Zhou_2019_CVPR} and translation, with the reference frame as the first frame.
\end{itemize}
Our multi-modal dataset is represented as \( \mathbf{X} = \{\mathbf{X}^{\text{rgb}}, \mathbf{X}^{\text{depth}}, 
% \mathbf{X}^{\text{hand}}, 
\mathbf{X}^{\text{gaze}}, \mathbf{X}^{\text{cam}} \} \), where modalities can be missing for each sub-dataset according to Tab.~\ref{tab:datasets}.

\begin{figure*}[t]
    \centering
    % \setlength{\tabcolsep}{0.0130\linewidth}
    %\begin{tabular}{@{}cc@{}}
    \includegraphics[width=\linewidth]{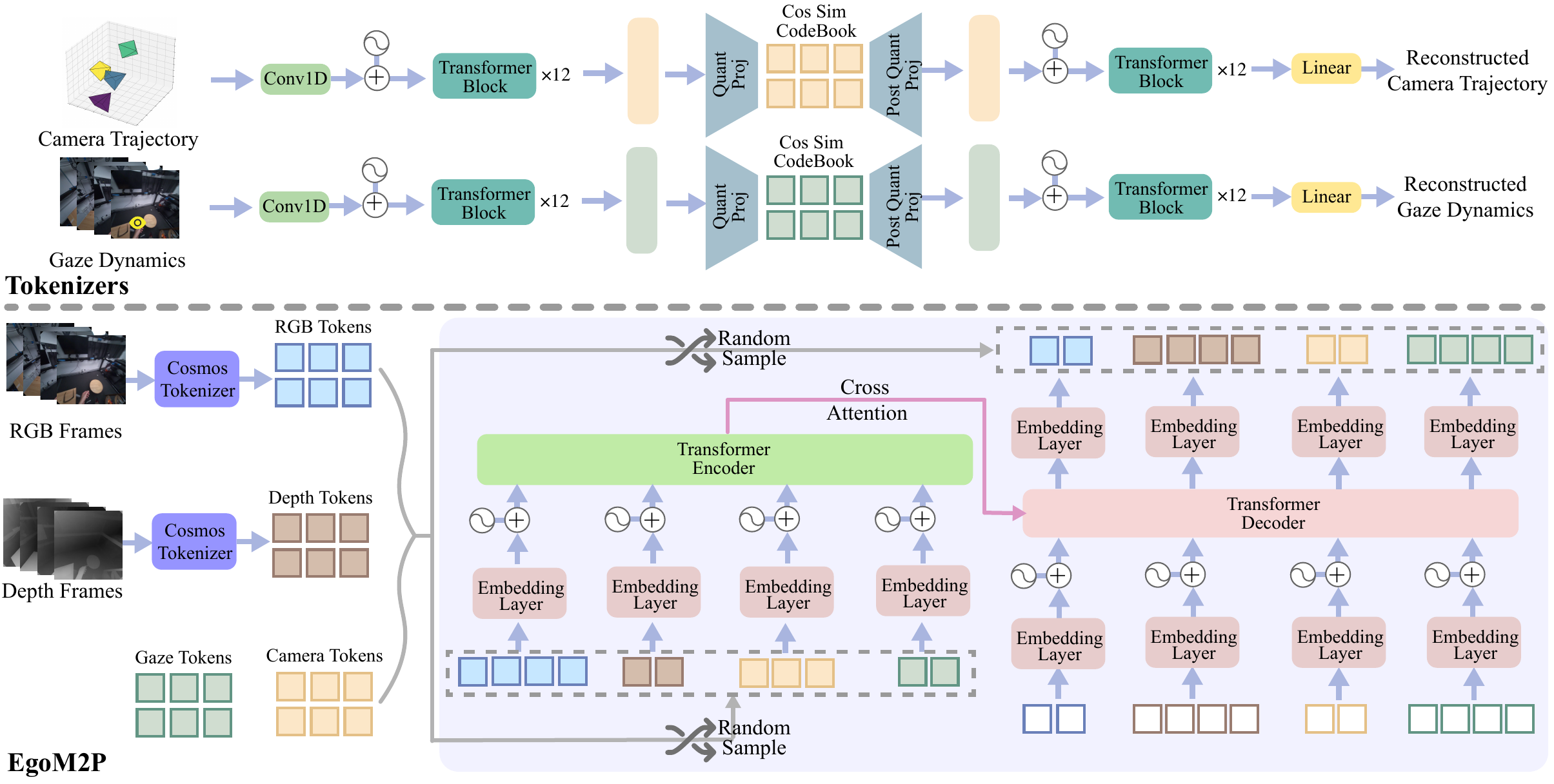}
    % \fbox{\rule{0pt}{2.5in} \rule{.9\linewidth}{0pt}}    %\end{tabular}

    \caption{\textbf{Network Architecture:} (1) We train VQ-VAE~\cite{Oord2017NeuralDR} tokenizers for camera trajectories and gaze dynamics (Sec. \ref{sec:tokenizer}), and adopt Cosmos tokenizers~\cite{agarwal2025cosmos} to tokenize RGB and depth streams. High-dimensional input modalities, including videos, gaze dynamics, and camera trajectories, are compressed into discrete tokens to serve as our training database. (2) Our \methodname~follows the architecture of T5-Base~\cite{10.5555/3455716.3455856}. We perform multimodal masked pretraining (Sec. \ref{sec:mask_pretrain}), where we randomly sample a fixed number of input and target tokens from our token database without overlap. For simplicity, we only visualize four frames here.\vspace{-3mm}} 
     % We use 1D positional embeddingfor camera and gaze and 3D for videos.
     %We design multimodal tokenizers with a unified architecture to compress multimodal data into compact discrete tokens.
\label{fig:network}
\end{figure*}

% For video tokenization \siwei{RGB or RGB-D?}, we leverage the SOTA Cosmos tokenizer \cite{agarwal2025cosmos}, 
We leverage the state-of-the-art (SOTA) Cosmos tokenizer \cite{agarwal2025cosmos} to tokenize video modalities,
applying a temporal compression rate of 4 and a spatial compression rate of 8 to convert the video into discrete tokens.
For other modalities, we train modality-specific tokenizers. To ensure adaptability when incorporating new modalities, we employ a unified tokenizer architecture using a Transformer-based vector quantized variational autoencoder (VQ-VAE)~\cite{Oord2017NeuralDR}. 
The network architecture is illustrated in the upper part of Fig.~\ref{fig:network}. 
For each modality \(\textit{mod} \in \{\text{gaze, cam}\}\), let \(N = \dim(\mathbf{X^{\textit{mod}}}) - 1\). The encoder \(\mathcal{E}\) begins by performing an \(N\)-dimensional convolution, represented as \(\text{Conv}_N\text{d}(\mathbf{X^{\textit{mod}}})\). This operation downsamples along the temporal axis by a factor of 2.
% and, for the hand modality, additionally downsamples the spatial axis by a factor of 3.
Next, it adds an $N$-dimensional positional embedding before passing the embedded data through 12 Transformer blocks. Each Transformer block attends to every pair-wise interaction among all tokens. % (spatio-)temporal tokens.
Then, we quantize the embeddings using the quantizer $\mathcal{Q}$ to learn modality-specific codebooks, employing cosine similarity as the distance metric, following \cite{yu2022vectorquantized, 4m}.
Finally, discrete codes are decoded through the decoder \(\mathcal{D}\), which mirrors the architecture of the encoder:
\begin{align*}
    \mathbf{z} &= \mathcal{E} (\mathbf{X}^{\textit{mod}}) \\
    \mathbf{q} &= \mathcal{Q} (\mathbf{z}) \\
    \hat{\mathbf{X}}^{\textit{mod}} &= \mathcal{D} (\mathbf{q})
\end{align*}

The overall training loss is:
\[
\mathcal{L_\textit{tok}} = \|\mathbf{X}^\textit{mod} - \hat{\mathbf{X}}^\textit{mod}\|_2^2 + \|\operatorname{sg}[\mathbf{z}] - \mathbf{q}\|_2^2 + \beta\,\|\mathbf{z} - \operatorname{sg}[\mathbf{q}]\|_2^2,
\]
where \(\operatorname{sg}[\cdot]\) is the stop-gradient operator and \(\beta\) balances the commitment loss. 
Due to hardware constraints, certain modalities, such as gaze, may contain invalid data when tracking is lost. We mask out invalid data and use masked L2 loss instead. See Supp. Mat. Sec. A.1 and C. for details.

% \[
% \text{Masked L2 Loss} = \frac{1}{\sum \text{mask}} \sum_{i=1}^{N} \text{mask}_i \times (y_i - \hat{y}_i)^2
% \] 

\subsection{Multimodal Masked Pretraining}
\label{sec:mask_pretrain}

4M~\cite{4m,4m21} introduces the Massively Multimodal Masked Modeling training scheme for static image modalities, where a small batch of sampled multimodal target tokens is predicted using another batch of sampled multimodal input tokens. However, when applying masked modeling to egocentric videos, 4M encounters several challenges: 1) There are no mechanisms to ensure temporal consistency. 2) The number of tokens per video sample is significantly larger than in image modalities, hindering efficient training and scalability. 3) The ratio of tokens across different modalities is highly imbalanced; \eg, each sample contains 170 times more video tokens than gaze tokens, which may lead to the neglect of critical information from less represented modalities. 4) Missing annotations in egocentric datasets can pose challenges, whereas 4M pseudo-labeled all modalities.

\noindent\textbf{Multimodal Token and Dataset Balancing.} In Sec.~\ref{sec:tokenizer}, we leverage our Transformer-based VQ-VAE and Cosmos Tokenizer \cite{agarwal2025cosmos} to tokenize each modality into temporally-aware discrete tokens. The large number of video tokens poses challenges for multimodal token pretraining. 
While aggressively compressing video tokens during tokenization might be beneficial, it can negatively impact video quality.
To address this, we downsample videos to 8 FPS, reducing the token count per video to 1/3 of its original amount.
After tokenization, the training set comprises roughly 4 billion multimodal tokens, whereas there are just 13 million gaze tokens.
% As datasets with precise hand annotations are smaller by two orders of magnitude in Tab.~\ref{tab:datasets}, video token counts are still imbalanced.
In addition, the scale of different datasets is also highly imbalanced, \eg, EgoExo4D \cite{Grauman2023EgoExo4DUS} has 160 times more samples than H2O \cite{Kwon_2021_ICCV}.
Training directly with these imbalanced datasets can cause the model to ignore modalities that have fewer tokens. Additionally, some datasets might suffer from overfitting, while others remain underfitted.
To mitigate this issue, we experiment with different sampling weights for both dataset sampling and token sampling across multi-modalities within the datasets. We discover that initially sampling a dataset with a probability proportional to its size, followed by sampling tokens from its modalities with uniform concentration parameters following 4M, results in the most stable training and optimal performance. See Supp. Mat. Sec. A.2 for details.

% we adopt a two-stage pretraining scheme. In the first stage, we sample data exclusively from larger-scale datasets~\cite{HoloAssist2023, Grauman2023EgoExo4DUS, li2024egogen}. \gen{After converging, we finetune it with the remaining smaller-scale datasets. To avoid catastrophic forgetting \cite{NEURIPS2019_fa7cdfad}, datasets used in the first stage are sampled with a small probability during finetuning.}
% The probability of sampling from each dataset is proportional to its size to ensure that training does not overfit one dataset while leaving others underfitted.

\noindent\textbf{Temporal Multimodal Token Embedding.} As shown in the lower part of Fig. \ref{fig:network}, for each modality, we use modality-specific embedding layers to map input tokens into a high-dimensional unified space, facilitating the alignment and integration of multimodal information. We then add sine-cosine positional embeddings, using 1D for gaze and camera tokens and 3D for video tokens.
The same approach is applied to target tokens.
Similar to 4M \cite{4m, 4m21}, we incorporate a learnable modality category embedding, which is shared for both input and target token embedding modules.

\noindent\textbf{Masking.} Masked modeling has demonstrated its efficacy in prior works \cite{He2021MaskedAA, tong2022videomae, 4m, 4m21, bachmann2022multimae}. 
4M requires aligned multimodal annotations and resorts to pseudo-labeling.
However, it is not practical to pseudo-label all modalities for egocentric datasets.
We represent missing modalities as placeholders and mask them out. See Supp. Mat. Sec. A.4 for details on handling missing modalities.
Similar to 4M, we reduce computational costs by applying input and target masking, encoding and decoding only a fixed number of visible tokens. While 4M caps this at 256, given that the number of tokens per video exceeds 5000, we increase this number to 2048 to accommodate more information. For each multimodal data sample $\mathbf{X}_i$, we sample how many visible tokens to use as inputs and targets for each available modality  \(\textit{mod} \in \{\text{rgb, depth, gaze, cam}\}\), then sample tokens in each clip $\mathbf{X}_i^{\textit{mod}}$ within these limits accordingly (See Supp. Sec. A.2). 
Visible input and target tokens are mutually exclusive.

\noindent\textbf{Model architecture.} 
Apart from the modality-specific embedding layers for the input and target tokens, the main architecture follows T5-Base \cite{10.5555/3455716.3455856}. 
The encoder applies self-attention to all sampled visible input tokens, integrating spatiotemporal information from multiple modalities. The decoder input is formed by masking the sampled visible target tokens. 
The decoder applies cross-attention with the encoder output as context and performs masked self-attention—restricted to tokens within the same modality—to predict those masked target tokens, which ensures that the decoder generates coherent tokens within the same modality. The training loss is cross-entropy. For each modality $m$:
% \begin{align*}
%     \mathcal{L}_{\text{total}} = \sum_{m=1}^{M} \mathbf{1}_m \cdot \mathcal{L}_m, \mathcal{L}_m = - \sum_{t=1}^{T_m} \sum_{c=1}^{C_m} y_{t,c} \log(\hat{y}_{t,c}),
% \end{align*}
\vspace{-4mm}
% For each modality m:
\[
\mathcal{L}_m = -\frac{1}{T_m} \sum_{t=1}^{T_m} \sum_{c=1}^{C_m} y_{t,c}^{(m)} \log\left(\hat{y}_{t,c}^{(m)}\right)
\]
\vspace{-1mm}
% Overall loss (averaging over modalities):
\[
\mathcal{L}_{\text{total}} = \frac{1}{|\mathcal{M}|} \sum_{m \in \mathcal{M}} \mathcal{L}_m
\]
% \[
% \mathcal{L}_{\text{token}} = -\frac{1}{\sum_{m \in \mathcal{M}} T_m} \sum_{m \in \mathcal{M}} \sum_{t=1}^{T_m} \sum_{c=1}^{C_m} y_{t,c}^{(m)} \log\left(\hat{y}_{t,c}^{(m)}\right)
% \]
where $\mathcal{M}$ is the set of modalities that have available tokens, 
% $\mathbf{1}_m$ denotes the availability of modalities, 
$T_m$ is the number of tokens in each modality, $C_m$ is the codebook size, and $\hat{y}_{t,c}$ is the predicted probability distribution for each token. See Supp. Mat. Sec. A.3 for details.

\subsection{Inference}
\label{sec:generation}
During training, we use variable masking rates to randomly mask out multimodal tokens that are encoded with temporal information.
Prior works \cite{4m} have shown that masked image models trained with this scheme function as order-agnostic autoregressive models
\cite{hoogeboom2022autoregressive}, allowing tokens to be decoded iteratively in random orders for parallel inference.
Similarly, we show that \methodname~is able to predict the distribution over masked tokens simultaneously, potentially providing the speed needed for real-time applications. As shown in Tab.~\ref{tab:tracking}, our method can predict the camera trajectory for a 60-frame video in 0.18 seconds (300+ FPS).

The parallel inference process can be formulated as a multi-step decoding procedure. 
We use a linear scheduling approach to predict $n$ target tokens over $s$ decoding steps. At each step, we randomly select $n/s$ target tokens and
first perform a forward pass of the pretrained model using all visible input tokens to predict the conditional distribution $\hat{y}_\text{cond}$ of selected target tokens in parallel. Next, we mask out all input tokens and perform a second forward pass to predict the unconditional probability $\hat{y}_\text{uncond}$ of selected target tokens using cross-attention on masked input tokens. The predicted target token distribution $\hat{y}$ is estimated using classifier-free guidance \cite{ho2021classifierfree} with weight $\omega$:
\vspace{-1mm}
\begin{align*}
    \hat{y} = (1 + \omega) \hat{y}_\text{cond} - \omega \hat{y}_\text{uncond}
\end{align*}
The final target token prediction is sampled from the predicted distribution $\hat{y}$ with nucleus sampling \cite{Holtzman2020The}. We find that for modalities with a small number of tokens, a single decoding step is sufficient. For video modalities, increasing decoding steps and predicting a subset of tokens at each step is beneficial. See Supp. Mat. Sec.~A.5 for pseudocode.

\subsection{Implementation Details}
For each multimodal clip, we set its length to 2 seconds. Non-video modalities have $T=60$ frames, while video modalities have $T=16$ frames due to the reduced FPS. The video resolution is 256$\times$256. 
After tokenization, the RGB and depth videos have 5120 tokens per sample, and the gaze and camera trajectory have 30 tokens per sample. See more details in Supp. Mat. Sec. A.

% For tokenizers, the codebook size is 64000 for video modalities, while the gaze and camera modalities have a codebook size of 256. The number of parameters for gaze and camera tokenizers is approximately 180 million. The camera trajectory tokenizer trains in 1 day, while the gaze tokenizer takes 12 hours, both using 4 NVIDIA GH200 superchips, each with an H100 GPU and 96GB of RAM.

% For \methodname, the total number of parameters is approximately 400 million. During training, we first randomly sample a dataset with probability proportional to its size; then, similar to 4M, we use Dirichlet distribution to sample visible input tokens and target tokens from available multimodalities. The maximum number of sampled tokens for both input and target is 2048. The total number of training tokens is 400 billion, randomly sampled and masked from our database of 4 billion tokens. The model training takes 15 hours using 256 NVIDIA GH200 superchips.  All networks are trained from scratch without initializing parameters from existing large models.

\section{Experiments}
We benchmark \methodname's multitasking abilities with SOTA models in downstream tasks, including egocentric perception and synthesis. We also benchmark it on unseen datasets without any fine-tuning to show the strong generalization ability of the pretrained feature. Additionally, we show \methodname~can be easily fine-tuned via post-training.

% \subsection{Tokenizer Evaluation}

% \gen{our rgb tokenizer vs. cosmos tokenizer on low FPS, resolution data. use egoexo4d eval set or holoassist}

% \noindent\TODO{camera tokenizer vs. chatcam? but there is no code available. needs confirmation. need to search for some existing hand and gaze tokenizers if they exist.} 

% \yiqian{For  gaze tokenizer , current I only find \url{https://arxiv.org/pdf/2406.08379}, it is a pre-print paper and no code available.}  
 
% \yiqian{For hand tokenizer, \url{https://github.com/LAHAproject/InAViT} releases the code, they use the tokens cropped from the image tokens using  the bounding box of the hand. \url{https://arxiv.org/pdf/2405.20030}, no code.}

% \subsection{Ablation Studies}

% \gen{larger model results? 960M 1B? if worse should we write?}

% \noindent\textbf{Multi-dataset Balancing.}

% \noindent\textbf{Visible Token}

\iffalse
\subsection{Egocentric Hand Motion Reconstruction}

We evaluate hand motion reconstruction from egocentric videos on the HOT3D dataset~\cite{banerjee2024introducing}.
We compare our method with the baseline HaMeR~\cite{pavlakos2024reconstructing}.
We report the mean per joint position error in hand-centric canonicalized coordinate systems.
HaMeR achieves an error of 6.05 cm when canonicalizing hand poses on a per-frame basis. However, when temporal smoothness is incorporated by canonicalizing hand motions relative to the first visible hand, the error increases significantly to 1.74 meters. This substantial rise in error is attributed to the highly fluctuating hand translations in HaMeR's predictions.
% , highlighting challenges in maintaining temporal consistency.
\fi

\subsection{Egocentric Camera Tracking}
\begin{table}[t]
    \centering
    \setlength{\tabcolsep}{3.5pt} % Reduce column spacing
    \renewcommand{\arraystretch}{1.} % Adjust row spacing
    \scriptsize
    \begin{tabular}{l|ccc|ccc|c}
        \toprule
        \multirow{2}{*}{Method} & \multicolumn{3}{c|}{EgoExo4D~\cite{Grauman2023EgoExo4DUS}} & \multicolumn{3}{c|}{ADT~\cite{pan2023ariadigitaltwinnew} (\textit{unseen})} &  \\
        &ATE$\downarrow$&RTE$\downarrow$&RRE$\downarrow$& ATE$\downarrow$&RTE$\downarrow$&RRE$\downarrow$& Time$\downarrow$ \\
        \midrule
        DROID-SLAM~\cite{teed2021droidslam} & 0.018 & 0.005 & 0.506 & 0.034 & 0.010 & 0.316 & 2.7s \\
        ACE-Zero~\cite{brachmann2024acezero} & 0.028 & 0.007 & 0.672  & 0.049 & 0.011 & 0.333 & 426s \\
        % Align3R (resize,DepthAnythingv2)  & 0.005 & 0.003 & 0.235 & 0.27 & ? & ?& ? & ? & ? \\
        Align3R~\cite{lu2024align3r}  & 0.019 & 0.006 & 0.762  &\textbf{0.028} & 0.010 &\textbf{0.276} & 372s \\ %(resize,DepthPro)
        % CasualSAM & ? & ? & ? & ? & ? & ?& ? & ? & ? \\ (1.6m)
        % MonST3R & ? & ? & ? & ? & ? & ?& ? & ? & ? \\
        % COLMAP & ? & ? & ? & ? & ? & ?& ? & ? & ? \\
        % Mega-SAM & ? & ? & ? & ? & ? & ?& ? & ? & ? \\
        \midrule
        % EgoMLVM-v1& 0.022 & 0.004 & 0.490  & 0.042 &0.008 & 0.548 & \textbf{0.1s }\\ % For ADT, EgoExo, does not right-multiply the GT pose
        % EgoMLVM e69-s1& 0.020 & \textbf{0.004} & \textbf{0.470}  & 0.034 &  \textbf{0.006} & 0.505 & \textbf{0.1s }\\ % EgoExo w/o 90 rot
        %MegaSAM
        
        % \multirow{2}{*}{EgoMLVM-e118-s2} & \multirow{2}{*}{\textbf{0.018}} & \multirow{2}{*}{\textbf{0.004}} & \multirow{2}{*}{\textbf{0.439}} & 0.033 & \textbf{0.006} & 0.483 & \multirow{2}{*}{\textbf{0.18s}}\\
        %  &  & & & - & - & - &  \\
        
         % -e115-s2
         % \multirow{2}{*}{EgoMLVM} & \multirow{2}{*}{\textbf{0.018}} & \multirow{2}{*}{\textbf{0.004}} & \multirow{2}{*}{\textbf{0.425}} & 0.033 & \textbf{0.006} & 0.487 & \multirow{2}{*}{\textbf{0.18s}}\\ &  & & & - & - & - &  \\
        % e92-s3
        \multirow{2}{*}{\methodname} & \multirow{2}{*}{\textbf{0.017}} & \multirow{2}{*}{\textbf{0.004}} & \multirow{2}{*}{\textbf{0.429}} & 0.032 & \textbf{0.006} & 0.490 & \multirow{2}{*}{\textbf{0.18s}}\\ &  & & & \textbf{\underline{0.026}} & \underline{\textbf{0.005}} & \underline{0.480} &  \\ % e1-s2 finetune
        
        \bottomrule
    \end{tabular}
    \caption{\textbf{Evaluation on camera tracking.} Compared to specialist SOTAs that require geometry test-time optimization, \methodname's feed-forward tracking results achieve comparable performance yet with significantly higher efficiency. We report the average runtime per sequence. \underline{Underlined} denotes post-training results (Sec. \ref{sec:post-training}).
    \vspace{-5mm}} %Note that all baselines require camera intrinsics to rectify the fish-eye images while our method can directly process the fish-eye images. 
    \label{tab:tracking}
\end{table}

\noindent\textbf{Evaluation Protocols.} We sample 200 video clips from the validation split of the EgoExo4D dataset~\cite{Grauman2023EgoExo4DUS} for evaluation. To assess \methodname's generalization to unseen dataset, we also evaluate on 200 video clips from the Aria Digital Twin (ADT) dataset~\cite{pan2023ariadigitaltwinnew}, which is entirely excluded from our training dataset. We process all input frames to 256$\times$256 resolution and 8 FPS. We use standard error metrics: Absolute Translation Error (ATE), Relative Translation Error (RTE), and Relative Rotation Error (RRE). 

\noindent\textbf{Baselines.} We compare with specialist camera tracking methods, including DROID-SLAM~\cite{teed2021droidslam}, ACE-Zero~\cite{brachmann2024acezero}, and Align3R~\cite{lu2024align3r}. Note that all baseline methods leverage explicit geometry constraints and perform bundle adjustment during test-time optimization. ACE-Zero and Align3R~\cite{lu2024align3r} also reply on off-the-shelf monocular depth and optical flow predictions as additional inputs. In contrast, \methodname~is trained \emph{without} any geometry modeling or 3D inductive bias and can predict camera poses directly from RGB inputs in a single feed-forward pass.

% The egocentric test images contain radial distortion, but all baselines can only process pinhole images. Therefore, we first rectify the distorted raw images using provided ground-truth camera intrinsics and feed the rectified images as inputs to baselines. For \methodname, we directly use the raw images as input. All input frames are spatially resized to 256$\times$256. 

\begin{figure}
    \centering
    \begin{subfigure}[b]{\linewidth}
    \centering
        \setlength{\tabcolsep}{0.0mm} 
        \renewcommand{\arraystretch}{0.0}
        \newcommand{\sz}{0.5}
    
        \begin{tabular}{cc}
            \includegraphics[width=\sz\linewidth]{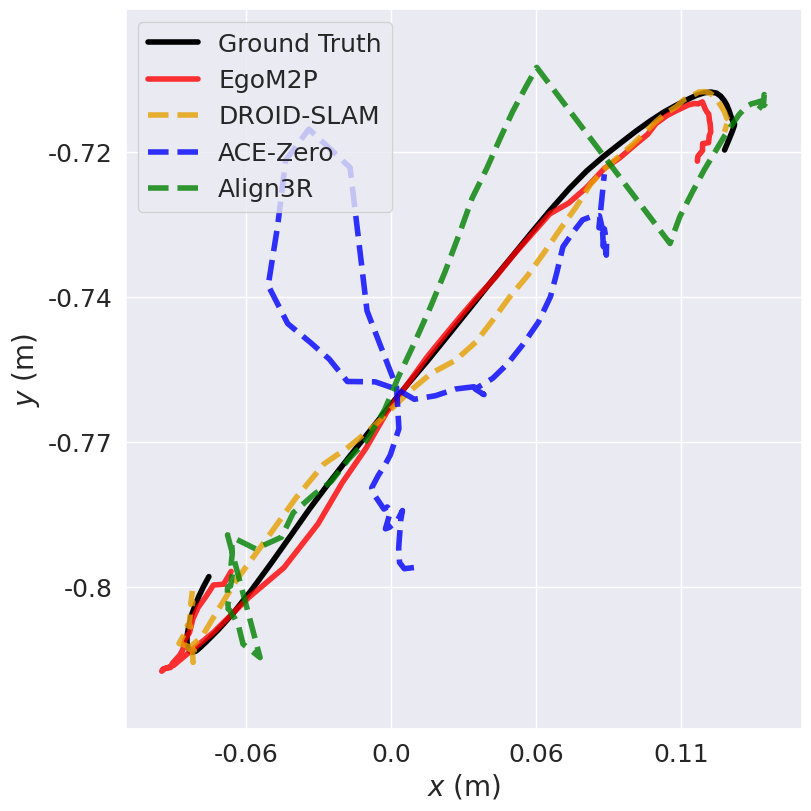}  & \includegraphics[width=\sz\linewidth]{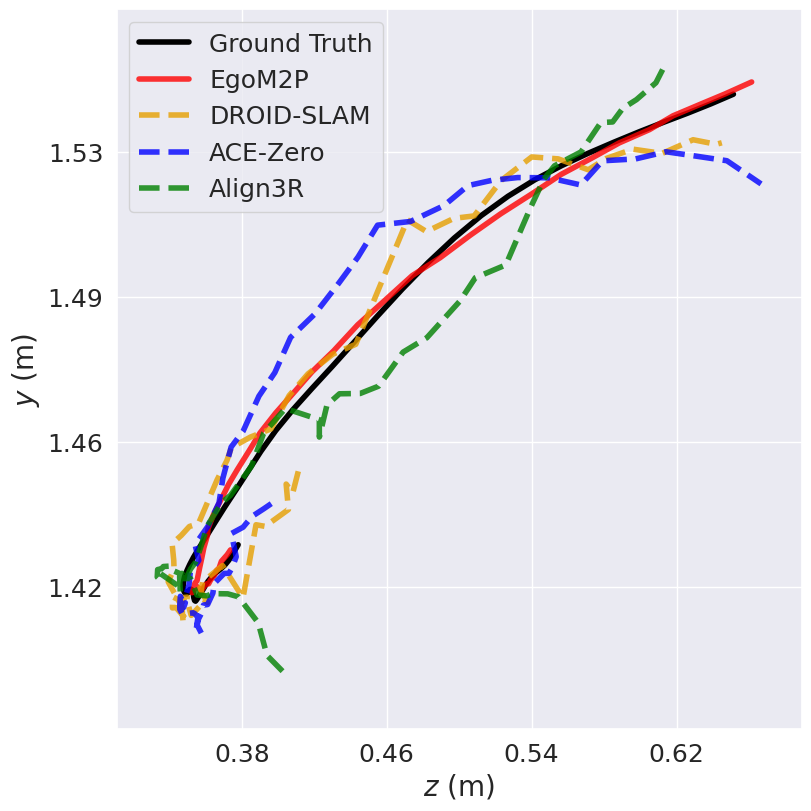} \\
            \text{\scriptsize{EgoExo4D~\cite{Grauman2023EgoExo4DUS}}} & \text{\scriptsize{ADT~\cite{pan2023ariadigitaltwinnew}}}
        \end{tabular}
        \caption{\textbf{Visualization of camera trajectories.} }
    \end{subfigure}
    \begin{subfigure}[b]{\linewidth}
    \centering
    \setlength{\tabcolsep}{0.0mm} 
    \tiny
    \renewcommand{\arraystretch}{0.0}
    \newcommand{\sz}{0.18}
    \newcommand{\raisei}{8}
    
    \begin{tabular}{cc cccc} 
    \phantom{+++} & Input & EgoM2P & DROID-SLAM~\cite{teed2021droidslam} & ACE-Zero~\cite{brachmann2024acezero} & Align3R~\cite{lu2024align3r} \\
    
    % Row for Frame 05
    \raisebox{8\height}{\rotatebox[origin=c]{90}{\makebox[0pt][c]{\scriptsize{Frame 05}}}} & 
    \includegraphics[width=\sz\linewidth]{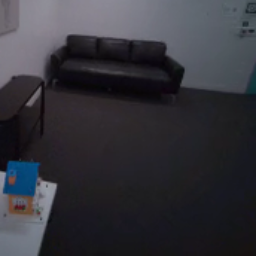} &
    \includegraphics[width=\sz\linewidth]{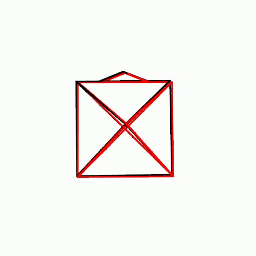} &
    \includegraphics[width=\sz\linewidth]{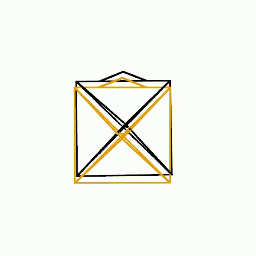} &
    \includegraphics[width=\sz\linewidth]{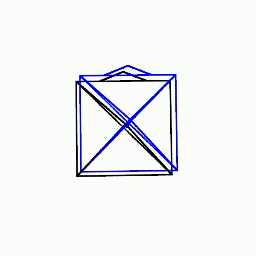} &
    \includegraphics[width=\sz\linewidth]{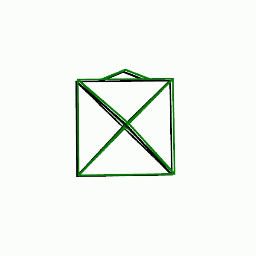} \\
    
    % Row for Frame 25
    \raisebox{8\height}{\rotatebox[origin=c]{90}{\makebox[0pt][c]{\scriptsize{Frame 25}}}} & 
    \includegraphics[width=\sz\linewidth]{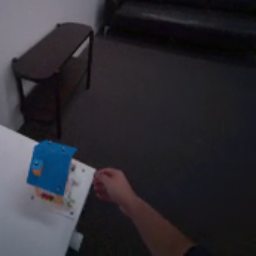} &
    \includegraphics[width=\sz\linewidth]{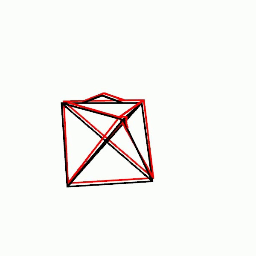} &
    \includegraphics[width=\sz\linewidth]{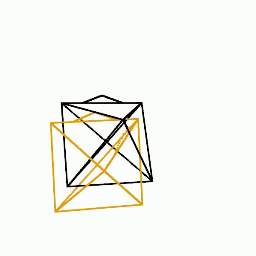} &
    \includegraphics[width=\sz\linewidth]{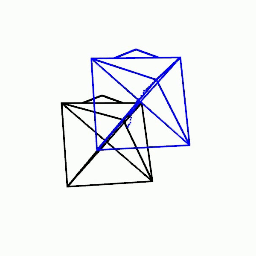} &
    \includegraphics[width=\sz\linewidth]{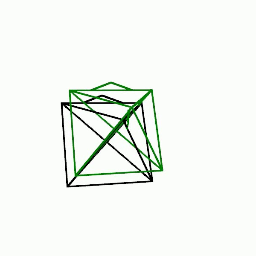} \\
    
    % Row for Frame 45
    \raisebox{8\height}{\rotatebox[origin=c]{90}{\makebox[0pt][c]{\scriptsize{Frame 45}}}} & 
    \includegraphics[width=\sz\linewidth]{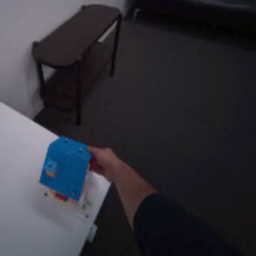} &
    \includegraphics[width=\sz\linewidth]{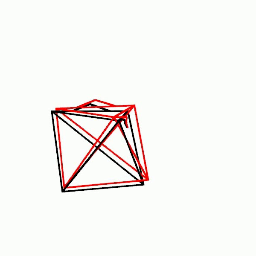} &
    \includegraphics[width=\sz\linewidth]{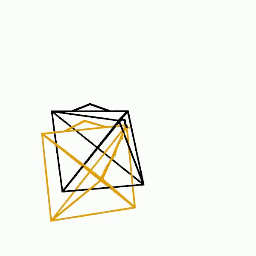} &
    \includegraphics[width=\sz\linewidth]{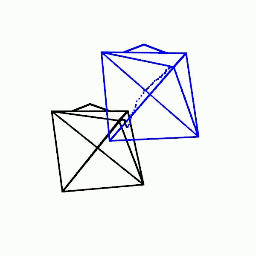} &
    \includegraphics[width=\sz\linewidth]{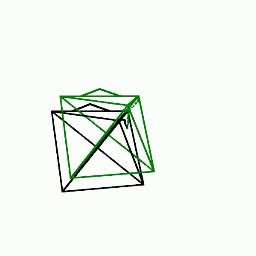} \\
    
    \end{tabular}
    \caption{\textbf{Comparison of camera tracking on ADT~\cite{pan2023ariadigitaltwinnew}}. Ground truth and predictions are represented by black and colored wireframes, respectively. }
    \end{subfigure}
    \caption{Egocentric capture often involves rapid head rotations, which challenges baseline tracking methods. However, \methodname~effectively predicts smooth and plausible camera poses in the shown examples. This capability also generalizes to the unseen ADT~\cite{pan2023ariadigitaltwinnew} dataset \textit{without} post-training.
    \vspace{-3mm}
    }
    \label{fig:rgb2pose}
\end{figure}

\noindent\textbf{Results.} See metrics in Tab.~\ref{tab:tracking}. Compared with specialist SOTA models that involve explicit geometry modeling, our versatile model trained without any 3D inductive bias achieves comparable performance. Notably, while all baselines require time-consuming test-time optimization, \methodname~achieves 300+ FPS inference speed, thanks to our parallel decoding approach. Egocentric camera tracking can sometimes be challenging due to rapid motion speed and little camera parallax. As shown in Fig.~\ref{fig:rgb2pose}, while baselines may suffer from temporal jitter and error accumulation in some cases, \methodname~can predict smooth and plausible camera trajectories in the sense that it learns to capture the uniqueness of egocentric motion. This capability can even generalize to the out-of-domain ADT test set.

\subsection{Egocentric Gaze Dynamics Estimation} 

\noindent\textbf{Evaluation Protocols.} 
We sample 1,000 videos from the validation split of the EgoExo4D dataset \cite{Grauman2023EgoExo4DUS}. We normalize both ground truth and predicted labels of all methods to the range $[0,1]$ and evaluate gaze estimation accuracy using the mean squared error (MSE).

\noindent\textbf{Baselines.} 
We compare \methodname~with two SOTA methods that predict 2D gaze locations from egocentric video: Huang et al. \cite{egocentric_gaze_prediction_huang} and Lai et al. \cite{GLC}, using their official implementations for evaluation.

\begin{figure}[t]
  \centering
  \includegraphics[width=1.0\linewidth]{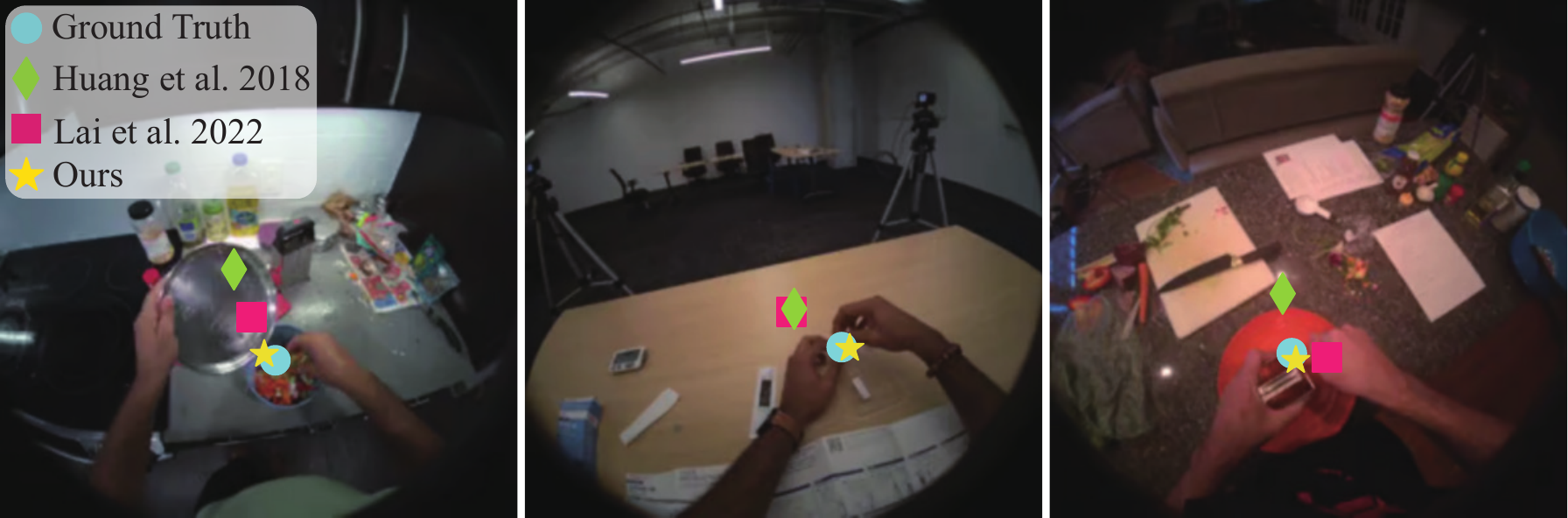}
  \caption{\textbf{Gaze dynamics estimation.} \methodname~can predict results that are more aligned with human intentions. \vspace{-3mm}
}
  \label{fig:gaze_estim}
\end{figure}

\noindent\textbf{Results.}
We use input videos at 30 FPS for the baselines, and following baselines' respective settings, input video frames are resized to 224$\times$224 for \cite{egocentric_gaze_prediction_huang} and 256$\times$256 for \cite{GLC}.
Our method achieves the lowest MSE (\textbf{0.0162}), outperforming Huang et al. \cite{egocentric_gaze_prediction_huang} (0.0255) and Lai et al. \cite{GLC} (0.0175). 
For qualitative comparisons, refer to Fig.~\ref{fig:gaze_estim} and Sup. Vid.
\methodname~produces more consistent gaze predictions, highlighting our capability to understand human intentions as one of its multitasking abilities.

% We compare \methodname~with two SOTA specialized methods: Huang et al. \cite{egocentric_gaze_prediction_huang} and Lai et al. \cite{GLC}, using their official implementations for evaluation. We sample 1,000 videos from the validation split of the EgoExo4D dataset \cite{Grauman2023EgoExo4DUS} for benchmarking.
% %
% We use input videos at 30 FPS for the baselines, and following their respective settings, input video frames are resized to 224$\times$224 for \cite{egocentric_gaze_prediction_huang} and 256$\times$256 for \cite{GLC}.
% %
% For fair comparisons, we normalize both the ground truth and predicted labels of all methods to the range $[0,1]$ and compute the mean squared error (MSE) loss for each approach.
% %
% Our method achieves the lowest MSE (0.0160), outperforming Huang et al. \cite{egocentric_gaze_prediction_huang} (0.0255) and Lai et al. \cite{GLC} (0.0175). 
% %
% For qualitative comparisons, please refer to Fig.~\ref{fig:gaze_estim} and the supplementary video.
% \methodname~produces more consistent gaze predictions in egocentric videos, highlighting our capability to understand human intentions as one of its multitasking abilities.

\subsection{Egocentric Monocular Video Depth Estimation}

\iffalse
\begin{table}[t]
    \centering
    \setlength{\tabcolsep}{0.8pt} % Reduce column spacing
    \renewcommand{\arraystretch}{1.} % Adjust row spacing
    \scriptsize
    \begin{tabular}{l|cc|cc|cc|c}
        \toprule
        \multirow{3}{*}{Method}
        & \multicolumn{2}{c|}{H2O~\cite{Kwon_2021_ICCV}} 
        & \multicolumn{4}{c|}{HOI4D~\cite{Liu_2022_CVPR}} 
        & \multirow{3}{*}{Time $\downarrow$} \\
        % \cmidrule{2-7}
        & \multicolumn{2}{c|}{} 
        & \multicolumn{2}{c|}{(\textit{unseen})} 
        & \multicolumn{2}{c|}{(\textit{fine-tune})} & \\
        & Abs Rel $\downarrow$ & $\delta_{1.25}\uparrow$ 
        & Abs Rel $\downarrow$ & $\delta_{1.25}\uparrow$ 
        & Abs Rel $\downarrow$ & $\delta_{1.25}\uparrow$ & \\
        \midrule
        RollingDepth~\cite{ke2024rollingdepth} & 0.087 & 90.5 & 0.057 & 97.6 & - & - & 37s \\ 
        Align3R~\cite{lu2024align3r} & 0.074 & 91.8 & \textbf{0.045} & \textbf{98.1} & - & - & 90s \\
        \midrule
        EgoMLVM & \textbf{0.054} & \textbf{95.8} & 0.064 & 97.7 & - & - & \textbf{1.9s} \\
        EgoMLVM +h2o & \textbf{0.053} & \textbf{96.2} & 0.063 & 97.9 & - & - & \textbf{1.9s} \\
        \bottomrule
    \end{tabular}
    \caption{\textbf{Evaluation on egocentric video depth estimation.} Compared with specialist SOTA baselines that require geometry-based test-time optimization, the versatile \methodname~achieves comparable performance while being significantly more efficient.
    \vspace{-3mm}}
    \label{tab:monocular_depth}
\end{table}
\fi

\begin{table}[t]
    \centering
    \setlength{\tabcolsep}{3pt} % Reduce column spacing
    \renewcommand{\arraystretch}{1.} % Adjust row spacing
    \scriptsize
    \begin{tabular}{l|cc|cc|c}
        \toprule
        \multirow{2}{*}{Method}
        & \multicolumn{2}{c|}{H2O~\cite{Kwon_2021_ICCV}} & \multicolumn{2}{c|}{HOI4D~\cite{Liu_2022_CVPR} (\textit{unseen})} &  \\
        & Abs Rel $\downarrow$ & $\delta_{1.25}\uparrow$ &
         Abs Rel $\downarrow$ & $\delta_{1.25}\uparrow$ &
        Time $\downarrow$  \\
        \midrule
        % 4M-21~\cite{4m21} & 0.138 & 82.9 & 0.076 & 94.5 &\textbf{ 1.7s} \\
        %unmasked RollingDepth~\cite{ke2024rollingdepth}  & 0.153 & 82.1 & 0.072 & 94.9 & 37s \\ 
        % RollingDepth~\cite{ke2024rollingdepth}  & 0.084 & 91.0 & 0.057 & 97.6 & 37s \\ 
        RollingDepth~\cite{ke2024rollingdepth} & 0.087 & 90.5 & 0.057 & 97.6 & 37s \\
        %unmasked Align3R~\cite{lu2024align3r}  & \textbf{0.134} & 84.7 & \textbf{0.062 }& \textbf{95.4} & 90s  \\
         % Align3R~\cite{lu2024align3r}  & 0.070 & 92.6 &\textbf{ 0.045} & \textbf{98.1 }& 90s  \\
         Align3R~\cite{lu2024align3r} & 0.074 & 91.8 & \textbf{0.045} & \textbf{98.1} & 90s\\
        \midrule
        %unmasked (before denoising) EgoMLVM &  0.170 & \textbf{85.4} & 0.070 & \textbf{95.4} & \textbf{1.9s} \\
        % EgoMLVM &  \textbf{0.052} & \textbf{96.0} & 0.064?& 97.7 ? & \textbf{1.9s} \\
        % EgoMLVM & \textbf{0.054} & \textbf{95.8} & 0.064 & 97.7 & \textbf{1.9s} \\
        % EgoMLVM +h2o & \textbf{0.053} & \textbf{96.2} & 0.063 & 97.9 & \textbf{1.9s} \\
        % EgoMLVM-e115-s3
        % \multirow{2}{*}{EgoMLVM-e115-s3} & \multirow{2}{*}{\textbf{0.054}} & \multirow{2}{*}{\textbf{96.1}} & 0.062 & \textbf{98.1} & \multirow{2}{*}{\textbf{0.8s}}\\ &  & & - & - &  \\

        % e92-s3
        \multirow{2}{*}{\methodname} & \multirow{2}{*}{\textbf{0.055}} & \multirow{2}{*}{\textbf{96.0}} & 0.061 & 98.0 & \multirow{2}{*}{\textbf{0.8s}}\\ &  & & \textbf{\underline{0.041}} & \textbf{\underline{99.0}} &  \\ %post-train e98-s9
        \bottomrule
    \end{tabular}
    \caption{\textbf{Evaluation on egocentric video depth estimation.} Compared to specialist SOTAs requiring geometry-based test-time optimization, the versatile \methodname~achieves comparable performance while being significantly more efficient. With post-training described in Sec. \ref{sec:post-training}, \methodname~excels (see \underline{underlined results}). 
    \vspace{-3mm}}
    \label{tab:monocular_depth}
\end{table}
\noindent\textbf{Evaluation Protocols.} We evaluate on the test split of H2O~\cite{Kwon_2021_ICCV}, which contains 236 two-second video clips. To validate our method's cross-domain generalization capability, we also evaluate on 100 two-second video clips from HOI4D~\cite{Liu_2022_CVPR}, which is entirely unseen during \methodname's training. We temporally downsample each video clip to 8 FPS and resize it to 256$\times$256 as input. We align the estimated relative depth with the GT depth by a sequence-level scale and translation factor and evaluate the depth accuracy with absolute relative error (Abs Rel) and the percentage of predicted depths within a 1.25-factor of true depth ($\delta_{1.25}$).

\noindent\textbf{Baselines.} We compare \methodname~with two specialized video depth estimators, RollingDepth~\cite{ke2024rollingdepth}, and Align3R~\cite{lu2024align3r}. Both methods employ pretrained networks~\cite{dust3r,ke2023repurposing} to estimate monocular or pair-wise depth maps and then conduct hierarchical sequence-level optimization to temporally align per-frame depths.

\begin{figure}[t!]
    \centering
    \setlength{\tabcolsep}{0.0mm} 
    \renewcommand{\arraystretch}{0.0}
    \newcommand{\sz}{0.23}
    \newcommand{\isz}{0.5mm}
    \newcommand{\raisei}{14}
    \begin{tabular}{c cc cc} 
    
    \phantom{+} & \multicolumn{2}{c}{H2O~\cite{Kwon_2021_ICCV}} & \multicolumn{2}{c}{HOI4D~\cite{Liu_2022_CVPR}} \\

          % \rotatebox[origin=c]{90}{\makebox[0pt][c]{\footnotesize{Input}}}
      \raisebox{\raisei\height}{\rotatebox[origin=c]{90}{\makebox[0pt][c]{\footnotesize{Input}}}}& 
      \includegraphics[width=\sz\linewidth]{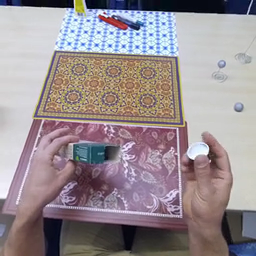} &
      \includegraphics[width=\sz\linewidth]{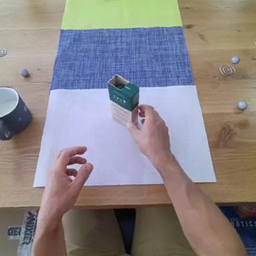} &  \hspace{\isz} 
      \includegraphics[width=\sz\linewidth]{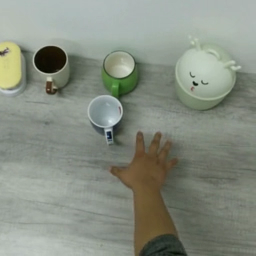} & 
     \includegraphics[width=\sz\linewidth]{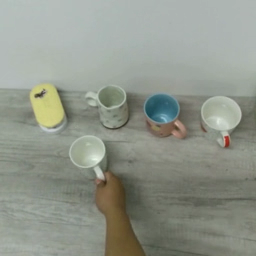} \\
    
      \raisebox{9\height}{\rotatebox[origin=c]{90}{\makebox[0pt][c]{\footnotesize{GT}}}} & 
      \includegraphics[width=\sz\linewidth]{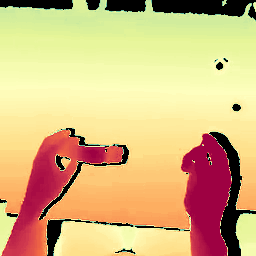} &  \includegraphics[width=\sz\linewidth]{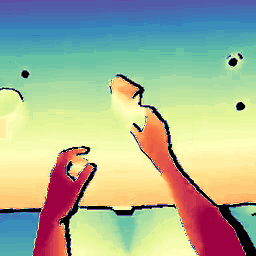} &  \hspace{1mm} 
      \includegraphics[width=\sz\linewidth]{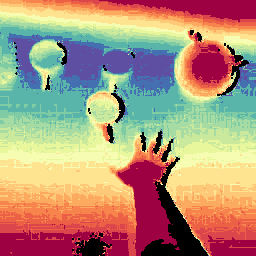} & 
      \includegraphics[width=\sz\linewidth]{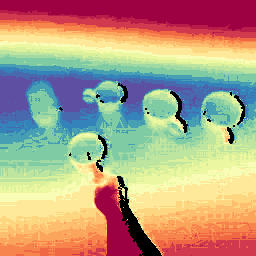} \\

      % \raisebox{\raisei\height}{\rotatebox[origin=c]{90}{\makebox[0pt][c]{\footnotesize{4M-21}~\cite{4m21}}}} & 
      % \includegraphics[width=\sz\linewidth]{img/rgb2depth/subject3_ego-o2-7_03_part1_frame12/4M-XL.png} &  \includegraphics[width=\sz\linewidth]{img/rgb2depth/subject3_ego-k2-7_01_part0_frame12/4M-XL.png} &  \hspace{\isz} 
      % \includegraphics[width=\sz\linewidth]{img/rgb2depth/H3-C2-N14-S215-s04-T5-08_frame8/4M-XL.png} & 
      % \includegraphics[width=\sz\linewidth]{img/rgb2depth/H3-C2-N47-S219-s02-T3-07_frame8/4M-XL.png} \\
      \raisebox{13\height}{\rotatebox[origin=c]{90}{\makebox[0pt][c]{\footnotesize{EgoM2P}}}} & 
      \includegraphics[width=\sz\linewidth]{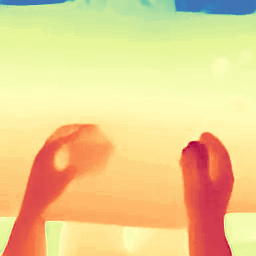} &  \includegraphics[width=\sz\linewidth]{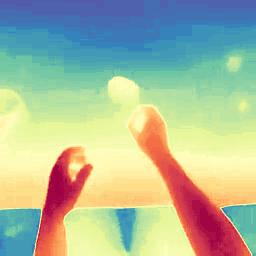} &  \hspace{\isz} 
      \includegraphics[width=\sz\linewidth]{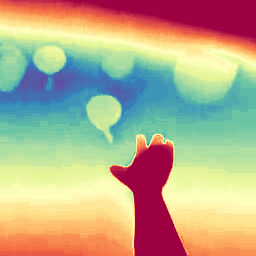} & 
      \includegraphics[width=\sz\linewidth]{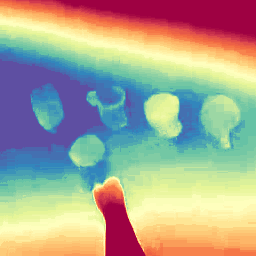}  \\

      \raisebox{\raisei\height}{\rotatebox[origin=c]{90}{\makebox[0pt][c]{\footnotesize{R-Depth}~\cite{ke2024rollingdepth}}}} & 
      \includegraphics[width=\sz\linewidth]{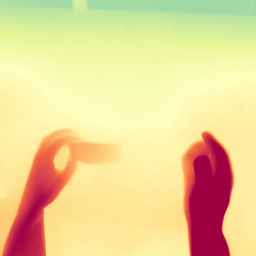} &  \includegraphics[width=\sz\linewidth]{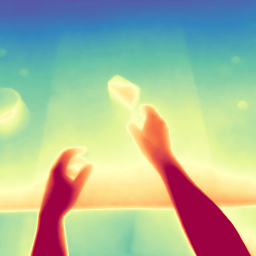} &  \hspace{\isz} 
      \includegraphics[width=\sz\linewidth]{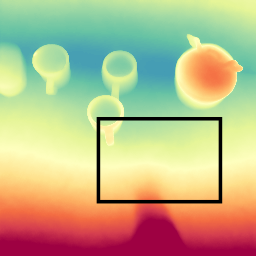} & 
      \includegraphics[width=\sz\linewidth]{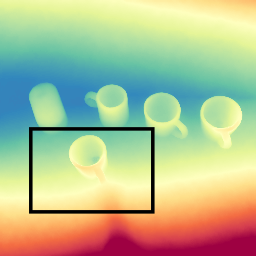} \\

      \raisebox{\raisei\height}{\rotatebox[origin=c]{90}{\makebox[0pt][c]{\footnotesize{Align3R}~\cite{lu2024align3r}}}} & 
      \includegraphics[width=\sz\linewidth]{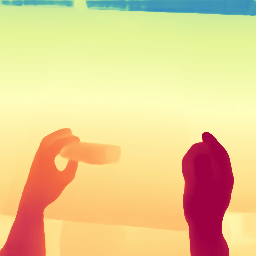} &  \includegraphics[width=\sz\linewidth]{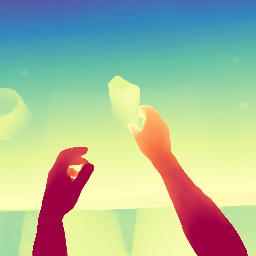} &  \hspace{\isz} 
      \includegraphics[width=\sz\linewidth]{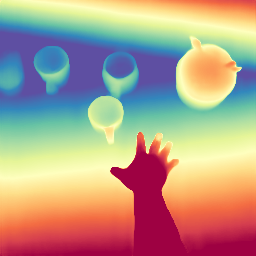} & 
      \includegraphics[width=\sz\linewidth]{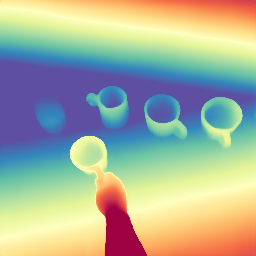} \\

    \end{tabular}
    \caption{\textbf{Egocentric video depth estimation}. \methodname~achieves comparable performance with specialist SOTA methods. Rolling Depth~\cite{ke2024rollingdepth} struggles in estimating the depth of hands, an important component in egocentric view, while our method can capture hand movement even in the out-of-domain HOI4D~\cite{Liu_2022_CVPR} dataset, \textit{without} any post-training or fine-tuning. \vspace{-3mm}}
    \label{fig:rgb2depth}
\end{figure}

\noindent\textbf{Results.} We report the results in Tab.~\ref{tab:monocular_depth} and provide visual comparisons in Fig.~\ref{fig:rgb2depth}. RollingDepth and Align3R require sequence-level optimization after per-frame predictions, which can take as long as one minute for a two-second sequence. In contrast, \methodname~predicts the entire depth video in a single end-to-end feed-forward pass, achieving at least 30x faster inference speed. Quantitatively, \methodname~achieves comparable performance with baseline methods and even the best $\delta_{1.25}$ on two test sets, particularly the unseen HOI4D dataset, which validates \methodname~'s generalization to out-of-domain egocentric data. Reducing the quantization error of the Cosmos tokenizer \cite{agarwal2025cosmos} would further improve our depth prediction result. Please refer to our supplementary videos for more visualization.

% \gen{add something like depth video prediction is also restricted by the quantization error of the cosmos tokenizer. maybe abs rel will be a bit high}

\subsection{Conditional Egocentric Video Synthesis}
\methodname~can synthesize RGB videos from other modalities. This section focuses on depth-to-RGB video synthesis. 

\noindent\textbf{Evaluation Protocols.} We randomly sampled 142 depth videos from the HoloAssist \cite{HoloAssist2023} test set and 100 depth videos from the ASE \cite{ase} dataset.  Note that the entire ASE dataset is unseen in our model's training and can be used to assess our method's generalizability.
For ASE, we use their labeled fisheye depth videos as inputs. For HoloAssist, since its depth labels are misaligned with the RGB frames, we employ RollingDepth \cite{ke2024rollingdepth} to generate pseudo-depth labels for each RGB frame as input. 
For quantitative metrics, we employ Fréchet Video Distance (FVD) \cite{fvd}, Structural Similarity Index (SSIM), Peak Signal-to-Noise Ratio (PSNR), and the perceptual metric LPIPS \cite{lpips}.

\noindent\textbf{Baselines.} We compare our method with two SOTA Con-trolNet-based approaches: Control-A-Video \cite{Control-A-Video} and ControlVideo \cite{ControlVideo}, both support depth as a conditional input.
% We compare two SOTA ControlNet-based video synthesis approaches: Control-A-Video \cite{Control-A-Video} and ControlVideo \cite{ControlVideo}, both support depth as conditional inputs.

\noindent\textbf{Results.} Since the two baselines require additional text input, we adopt different templates for each dataset. In HoloAssist, we use the template ``Two hands with \{object\}'', where \{object\} represents the item being held in the RGB video. For ASE, since it consists of indoor videos, we use the prompt ``Indoor scenes''.
In Fig.~\ref{fig:depth2rgb}, we present a qualitative comparison highlighting the challenges faced by two baselines in generating RGB frames that accurately correspond to the input depth maps.
These baseline methods often produce outputs with significant discrepancies in semantic and geometric alignment, especially on the ASE~\cite{ase} dataset. In contrast, our approach \methodname~demonstrates superior performance by maintaining alignment between the depth maps and the generated RGB frames.
% Specifically, when using ASE \cite{ase} depth as conditional input, the baseline methods frequently exhibit substantial hallucinations, resulting in outputs that do not align well with the depth information. 
% Our method, however, effectively synthesizes egocentric videos that are well-aligned with the provided depth inputs.
The quantitative results are reported in Tab.~\ref{tab:depth2rgb}. On the HoloAssist dataset \cite{HoloAssist2023}, our model outperforms the baseline approaches. 
On the ASE dataset \cite{ase}, which is unseen and stylistically different from our training data, our method demonstrates strong generalization capabilities, generating egocentric videos that more closely resemble real ones, as indicated by a lower FVD score. In contrast, baseline models, initialized with the powerful Stable Diffusion \cite{9878449}, tend to produce hallucinations. While they achieve higher PSNR scores, their outputs often deviate from the true egocentric video distribution, as indicated by their higher FVD scores.

\begin{table}[t]
    \centering
    \setlength{\tabcolsep}{0.8pt} % Reduce column spacing
    \renewcommand{\arraystretch}{1.} % Adjust row spacing
    \scriptsize
    \scalebox{0.92}{
    \begin{tabular}{l|cccc|cccc}
        \toprule
        \multirow{2}{*}{Method}
        & \multicolumn{4}{c|}{HoloAssist~\cite{HoloAssist2023} } & \multicolumn{4}{c}{ASE~\cite{ase} (\textit{unseen})}   \\
        & FVD$^*$$\downarrow$  & SSIM $\uparrow$  & PSNR $\uparrow$ &LPIPS $\downarrow$ &FVD$^*$$\downarrow$  & SSIM $\uparrow$  & PSNR $\uparrow$ &LPIPS $\downarrow$   \\
        \midrule
        Control-A-Video~\cite{Control-A-Video} & 2.309  & 0.185  &  9.25  &  0.677 &2.226  & 0.289  &  \textbf{11.11}  &  0.817 \\
        ControlVideo~\cite{ControlVideo}& 1.363 & 0.235  &  8.18  &  0.653 & 1.392  & 0.275  &  10.46  &  \textbf{0.676} \\ 
        % EgoMLVM & \textbf{0.759} & \textbf{0.592} & \textbf{15.26} & \textbf{0.339} & \textbf{0.767}  & \textbf{0.394}  &  8.73  &  0.679 \\ 
        % EgoMLVM-e92-s2 & \textbf{0.759} & \textbf{0.592} & \textbf{15.26} & \textbf{0.339} & \textbf{1.336}  & \textbf{0.308}  &  6.923  &  0.715 \\
        \midrule
        % -e92-s6 // -e92-s2
        \multirow{2}{*}{\methodname} & \multirow{2}{*}{\textbf{0.759}} & \multirow{2}{*}{\textbf{0.592}} &
        \multirow{2}{*}{\textbf{15.163}} &
        \multirow{2}{*}{\textbf{0.336}} & \textbf{1.336}  & \textbf{0.308}  &  6.923  &  0.715 \\
        & &  & & & \textbf{\underline{0.525}} & \textbf{\underline{0.594}} & \textbf{\underline{16.924}} & \textbf{\underline{0.520}} \\
        \bottomrule
    \end{tabular}
    }
    \caption{\textbf{Evaluation on depth-to-RGB video synthesis.} \textit{EgoM2P} outperforms baselines on the HoloAssist test set, producing higher-quality egocentric videos. On the unseen ASE dataset, it generates videos that more closely resemble real ones with a lower FVD$^*$ (FVD$/10^3$). With post-training (Sec. \ref{sec:post-training}), \methodname~excels on unseen datasets indicated by \underline{underlined results}.
    \vspace{-3.1mm}}
    \label{tab:depth2rgb}
\end{table}

% \begin{figure}[t]
%   \centering
%   \includegraphics[width=0.9\linewidth]{img/depth2rgb.pdf}
  
%   \caption{
% \textbf{Comparison of Depth-to-RGB Video Synthesis.} Red boxes highlight incorrectly generated fingers in baselines, while ours generates meaningful hand motion. Our results show improved alignment with the input depth, minimizing hallucinations.
% }
%   \label{fig:depth2rgb}
% \end{figure}

\begin{figure}[t!]
    \centering
    \setlength{\tabcolsep}{0.0mm} 
    \renewcommand{\arraystretch}{0.0}
    \newcommand{\sz}{0.24}
    \newcommand{\isz}{0.7mm}
    \newcommand{\raisei}{14}
    \begin{tabular}{c cc cc} 
    
    \phantom{+} &  \scriptsize{Input Depth} & 
    \fontsize{6.5pt}{\baselineskip}\selectfont  Control-A-Video\cite{Control-A-Video} & \scriptsize{ControlVideo}\cite{ControlVideo} & \scriptsize{Ours} \\

      \raisebox{\raisei\height}{\rotatebox[origin=c]{90}{\makebox[0pt][c]{}}}& 
      \includegraphics[width=\sz\linewidth]{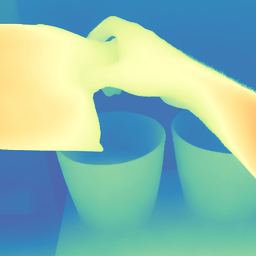} &
      \includegraphics[width=\sz\linewidth]{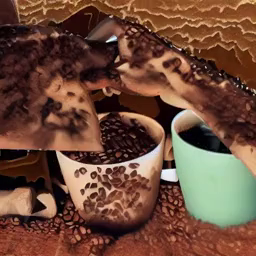} &  
      \includegraphics[width=\sz\linewidth]{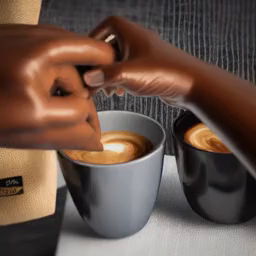} & 
     \includegraphics[width=\sz\linewidth]{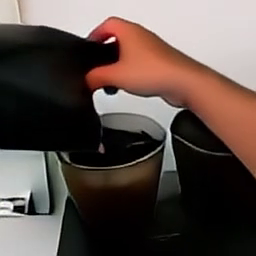} \\
    
      \raisebox{9\height}{\rotatebox[origin=c]{90}{\makebox[0pt][c]{\footnotesize{HoloAssist \cite{HoloAssist2023}}}}} & 
     \includegraphics[width=\sz\linewidth]{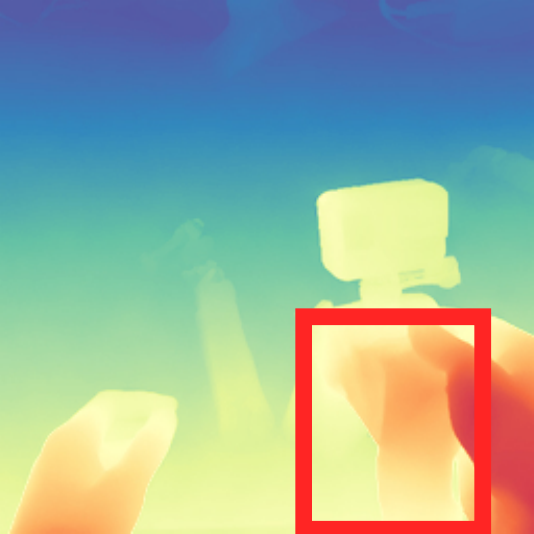} &
      \includegraphics[width=\sz\linewidth]{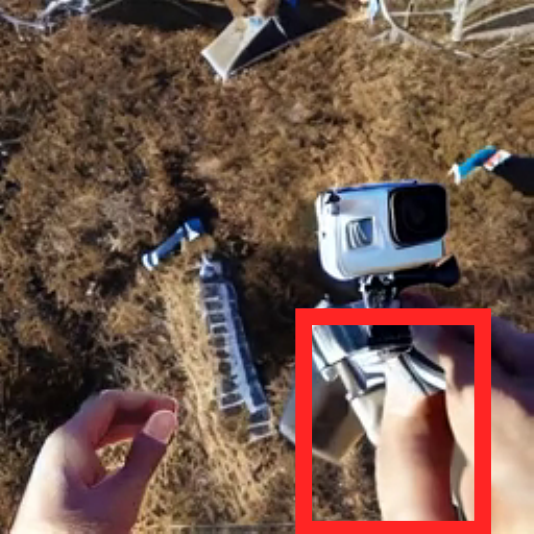} &  
      \includegraphics[width=\sz\linewidth]{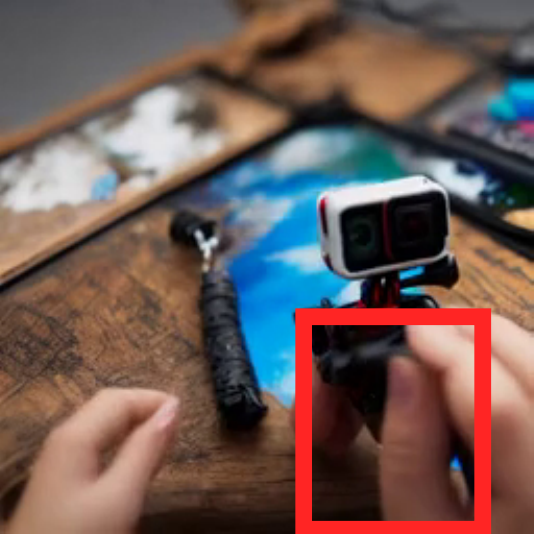} & 
     \includegraphics[width=\sz\linewidth]{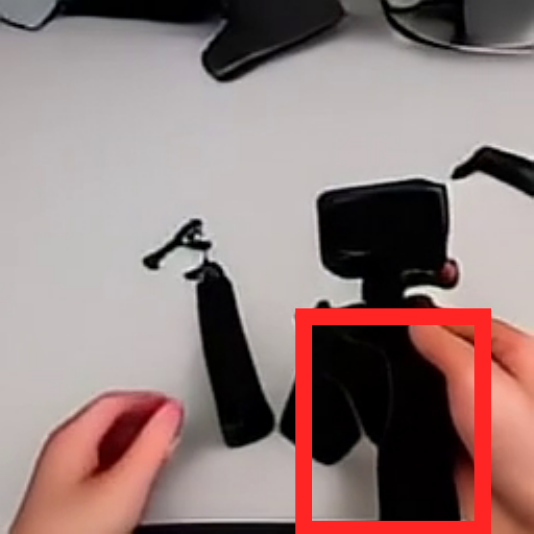} \\
    \vspace{\isz}
      \raisebox{\raisei\height}{\rotatebox[origin=c]{90}{\makebox[0pt][c]{}}} & 
      \includegraphics[width=\sz\linewidth]{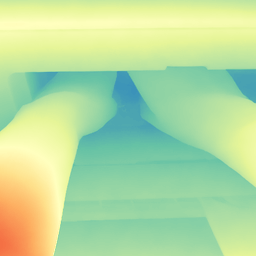} &
      \includegraphics[width=\sz\linewidth]{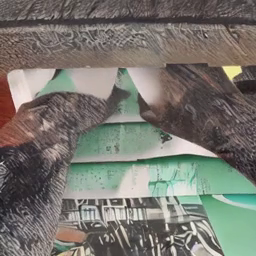} &  
      \includegraphics[width=\sz\linewidth]{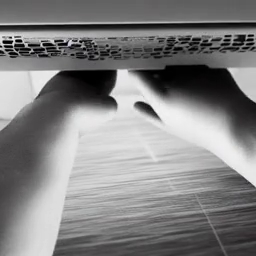} & 
     \includegraphics[width=\sz\linewidth]{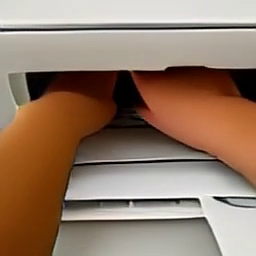}  \\
     
      \raisebox{\raisei\height}{\rotatebox[origin=c]{90}{\makebox[0pt][c]{}}} & 
      \includegraphics[width=\sz\linewidth]{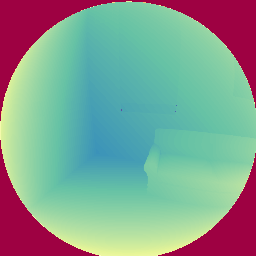} &  \includegraphics[width=\sz\linewidth]{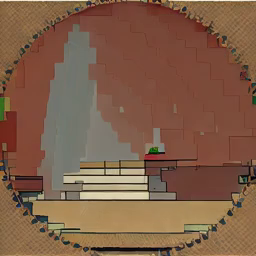} &   
      \includegraphics[width=\sz\linewidth]{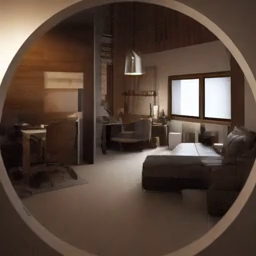} & 
      \includegraphics[width=\sz\linewidth]{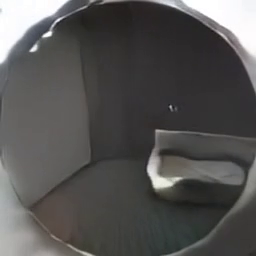} \\

      \raisebox{\raisei\height}{\rotatebox[origin=c]{90}{\makebox[0pt][c]{\footnotesize{ASE}~\cite{ase}}}} & 
      \includegraphics[width=\sz\linewidth]{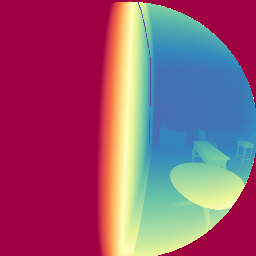} &  \includegraphics[width=\sz\linewidth]{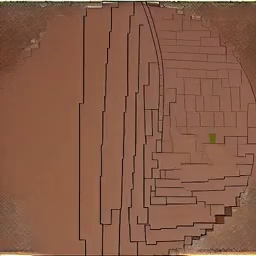} &   
      \includegraphics[width=\sz\linewidth]{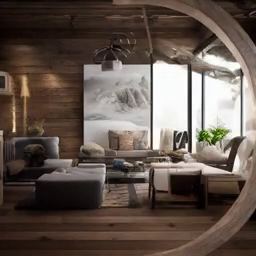} & 
      \includegraphics[width=\sz\linewidth]{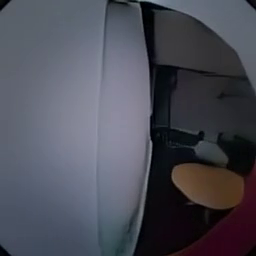} \\

      \raisebox{9\height}{\rotatebox[origin=c]{90}{\makebox[0pt][c]{}}} & 
      \includegraphics[width=\sz\linewidth]{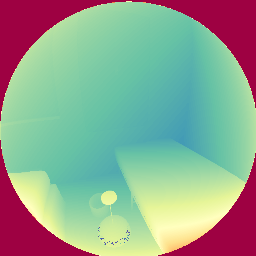} &  \includegraphics[width=\sz\linewidth]{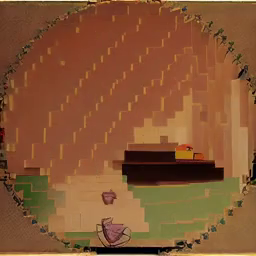} &   
      \includegraphics[width=\sz\linewidth]{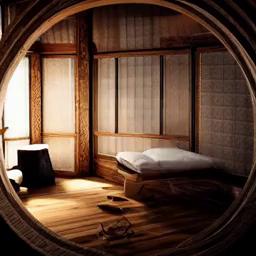} & 
      \includegraphics[width=\sz\linewidth]{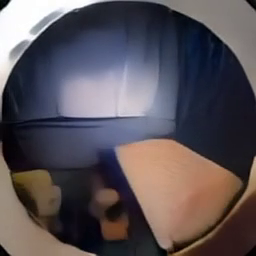}  \\
      
    \end{tabular}
    \caption{\textbf{Comparison of depth-to-RGB video synthesis.} Red boxes highlight incorrectly generated fingers in baselines, while ours generate meaningful hand motion. Our results show improved alignment with the input depth, minimizing hallucinations. \textit{No} post-training was applied for ASE results in this figure.\vspace{-3mm}}
    \label{fig:depth2rgb}
\end{figure}

% \noindent\textbf{Hand-to-RGB.}
% Evaluate on 118 videos, 4s, 8 FPS, 32frames. 
% For hand-to-RGB synthesis, we compare our method to the hand-mask-guided image synthesis approach CosHand \cite{CosHand}. CosHand utilizes hand masks as guiding signals, which we estimate using the Grounded SAM \cite{Grounded_SAM}. Utilizing the first frame as reference, CosHand generates RGB frames based on the estimated hand mask sequence from the original video.

% \url{https://excitedbutter.github.io/Instructional-Video-Generation/}

% \url{https://universal-simulator.github.io/unisim/}

% \noindent\textbf{Camera-to-RGB.}
% Evaluate on 1000 videos, 4s, 8 FPS, 32frames. 
% For camera-to-RGB synthesis, we generate RGB video starting from the first frame of the test video, guided by the camera trajectories of the original video. We compare our method with CamI2V \cite{CamI2V}, a camera-controlled image-to-video synthesis approach. We use video frames and cameras from the EgoExo4D \cite{Grauman2023EgoExo4DUS} dataset as input. 
% FAR

\subsection{Post-Training}
\label{sec:post-training}

The pretrained \methodname~demonstrates strong generalization abilities on cross-dataset generalization tests. Additionally, in Tab. \ref{tab:tracking}, \ref{tab:monocular_depth}, and \ref{tab:depth2rgb}, we show that \methodname~can be further enhanced via post-training on the training sets of unseen datasets, enabling rapid adaptation to new domains with minimal data. See more details in Supp. Mat. Sec. D.

% show \methodname~can via post-training on the training sets of those unseen datasets.

\section{Conclusion}

We propose \methodname, the first multimodal and multitask large egocentric model integrating four common modalities in egocentric vision. To handle complex spatiotemporal dynamics in multimodalities, we propose a unified temporal tokenizer architecture to tokenize gaze and camera trajectory into discrete tokens encoded with temporal information. To address heterogeneity in egocentric datasets, we extend multimodal masked modeling to the video domain and pretrain \methodname~with 400 billion tokens sampled from our 4 billion multimodal token database. 
\methodname~matches or surpasses state-of-the-art specialist models and demonstrates efficiency in various downstream applications, including egocentric camera tracking, gaze estimation in egocentric videos, egocentric monocular depth estimation, and conditional egocentric video synthesis.

\noindent\textbf{Limitations.} 
The visual quality of the synthesized videos is inherently limited by current state-of-the-art video tokenizers~\cite{agarwal2025cosmos}. Although better tokenizers could reduce video quality loss during quantization, this is not our main focus. 
Leveraging a diffusion decoder conditioned on discrete video tokens to enhance visual quality could be effective.

\noindent\textbf{Future Work.} Wearable devices have constrained computing resources and demand real-time processing for seamless human-computer interactions. We aim to explore performance optimizations for \methodname~on embedded GPUs. 
In this work, we consider the most common modalities in egocentric vision. Integrating hand motion, audio, text, etc., into the model is a promising direction for future work.
% Model distillation offers a promising avenue for improvement, aligning with ongoing advancements in Small Language Models \cite{abdin2024phi-}. 

\noindent\textbf{Acknowledgements.} 
Gen Li was supported by a Microsoft Spatial AI Zurich Lab PhD scholarship,
Yutong Chen was supported by the Swiss Innovation Agency Innosuisse, and Kaifeng Zhao was supported by the SDSC PhD fellowship. This work was also supported as part of the Swiss AI Initiative by a grant from the Swiss National Supercomputing Centre (CSCS) under project ID a03 on Alps. We sincerely thank Siwei Zhang for the fruitful discussions.

{
    \small
    \bibliographystyle{ieeenat_fullname}
    \bibliography{main}
}

\clearpage
\begingroup

\twocolumn[
\begin{center}
    {\Large \bf \Large{\bf EgoM2P: Egocentric Multimodal Multitask Pretraining} \\ -- Supplementary Material -- \par}
  \vspace*{30pt}
\end{center}
]

\appendix
\setcounter{page}{1}
\setcounter{table}{0}
\setcounter{algorithm}{0}
\setcounter{figure}{0}
\setcounter{equation}{0}
\counterwithin{figure}{section}
\counterwithin{table}{section}
\counterwithin{algorithm}{section}
\renewcommand{\thetable}{\thesection.\arabic{table}}
\renewcommand{\thefigure}{\thesection.\arabic{figure}}
\renewcommand{\theequation}{\thesection.\arabic{equation}}

\section{Implementation Details}

\subsection{Tokenizers}
\label{sup:tokenizer}
We tokenize spatiotemporal multimodalities into discrete tokens. For RGB and depth video, we use the state-of-the-art (SOTA) Cosmos tokenizer \cite{agarwal2025cosmos}. For gaze dynamics and camera trajectory, we train modality-specific tokenizers based on vector-quantized autoencoder.

For the gaze data $\mathbf{X}^{\text{gaze}} \in \mathbb{R}^{T \times 2}$, we first apply a convolution with a kernel size of 2 to temporally downsample it while mapping the channel dimension from 2 to 768. Then, we use 12 Transformer blocks (ViT-B) with self-attention to encode the data. Following best practices, we use cosine-sine similarity codebook with normalized codes. During training, we update the codebook entries to make sure all entries are used effectively. We track the exponential moving average of the codebook entry usage and replace under-utilized codes with the EMA dead code threshold. Quantized discrete tokens are then fed into the decoder with 12 Transformer blocks (ViT-B). Due to the hardware constraints, gaze data obtained from headsets, especially HoloLens \cite{HoloLens}, contains considerable invalid numbers. We choose not to discard these sequences, mask out invalid numbers, and use them as input. While calculating the reconstruction loss, we use masked L2 loss:
\[
\mathcal{L_{\text{gaze}}} = \frac{\sum_{i=1}^{N} m_i (y_i - \hat{y}_i)^2}{\sum_{i=1}^{N} m_i}
\]
where \( y_i \) is the ground truth, \( \hat{y}_i \) is the predicted value, and \( m_i \) is the binary mask indicating valid values (1 for valid, 0 for invalid). The denominator ensures normalization by the number of valid elements. By leveraging the smoothness of deep networks, invalid gaze could also be predicted.

For the camera trajectory data $\mathbf{X}^{\text{cam}} \in \mathbb{R}^{T\times9}$, we select the first two columns of the rotation matrix and the camera translation and stack them together. The only difference with the gaze tokenizer is the temporal convolution. This convolution performs temporal downsampling with a factor of 2 and maps the channel from 9 to 768. The training details are listed in Tab. \ref{tab:tokenizer_settings}. 
% Tokenizers are trained with 4 NVIDIA GH200 superchips with H100 GPUs. 
``Batch size'' refers to the number of samples per GPU. The ``Learning rate'' is determined by multiplying the ``Base Learning rate'' by the ``Total batch size'' and then dividing by 256 following~\cite{goyal2018accuratelargeminibatchsgd}.

The codebook size is 64000 for video modalities, while the gaze and camera modalities have a codebook size of 256. The number of parameters for gaze and camera tokenizers is approximately 180 million. The camera trajectory tokenizer trains in 1 day, while the gaze tokenizer takes 12 hours, both using 4 NVIDIA GH200 superchips, each with an H100 GPU and 96GB of RAM.

\begin{table}[t]
    \centering
    \begin{tabular}{l |c c}
        \toprule
        Configuration & Gaze Dyn. & Camera Traj. \\
        \midrule
        Codebook size & \multicolumn{2}{c}{256} \\
        Temporal compression & \multicolumn{2}{c}{2} \\
        Code latent dimension & \multicolumn{2}{c}{32} \\
        EMA dead code threshold & \multicolumn{2}{c}{2} \\
        Codebook EMA & \multicolumn{2}{c}{0.99} \\
        $l_2$-normalized codes \cite{yu2022vectorquantized} & \multicolumn{2}{c}{\checkmark} \\
        Codebook weight & \multicolumn{2}{c}{1.0} \\
        Commitment weight $\beta$ & \multicolumn{2}{c}{1.0} \\
        \midrule
        % Convolution & & \\
        Encoder architecture &  \multicolumn{2}{c}{ViT-B}\\
        Decoder architecture & \multicolumn{2}{c}{ViT-B} \\
        Loss function & Masked MSE & MSE \\
        \midrule
        Optimizer & \multicolumn{2}{c}{AdamW \cite{loshchilov2018decoupled}} \\
        Opt. momentum & \multicolumn{2}{c}{$\beta_1, \beta_2 = 0.9, 0.95$} \\
        Weight decay & \multicolumn{2}{c}{0.05} \\
        Base learning rate & 5e-5 & 2.5e-5\\
        Learning rate & 1e-4 & 5e-5\\
        Batch size &  \multicolumn{2}{c}{128} \\
        Total batch size &  \multicolumn{2}{c}{512} \\
        Max gradient norm & \multicolumn{2}{c}{1}  \\
        Learning rate sched. & \multicolumn{2}{c}{Cosine decay} \\
        Training epochs &   \multicolumn{2}{c}{200}   \\
        Warmup epochs & \multicolumn{2}{c}{5} \\
        Data type & \multicolumn{2}{c}{float32} \\
        \bottomrule
    \end{tabular}
    \caption{\textbf{Tokenizer training settings.}}
    \label{tab:tokenizer_settings}
\end{table}

\subsection{Multimodal Token Sampling}
\label{sup:sampling}
First, we randomly sample a dataset from eight curated egocentric datasets with probability proportional to the number of samples they contain. Then, we sample tokens from available multimodalities with a family of symmetric Dirichlet distributions:

1. \textbf{Sampling a Dirichlet Distribution:} Let $\boldsymbol{\alpha}_1$, $\boldsymbol{\alpha}_2$, $\boldsymbol{\alpha}_3$, $\boldsymbol{\alpha}_4$ be the four concentration parameters of the Dirichlet distributions. We select \(\boldsymbol{\alpha}_i\) with uniform probability:
   \begin{align*}
       \boldsymbol{\alpha}_1 &= (0.01, 0.01, 0.01, 0.01), \\
   \boldsymbol{\alpha}_2 &= (0.1, 0.1, 0.1, 0.1), \\
   \boldsymbol{\alpha}_3 &= (1, 1, 1, 1), \\
   \boldsymbol{\alpha}_4 &= (10, 10, 10, 10).
   \end{align*}
If the dataset contains missing modalities, the selected concentration parameter vector \(\boldsymbol{\alpha}_i\) is adjusted to include only the parameters corresponding to the available modalities.
   
2. \textbf{Sampling a Probability Vector:}  When \(\boldsymbol{\alpha}_i\) is sampled, we then sample a probability vector \(\boldsymbol{\theta}\) from the chosen Dirichlet distribution:
   \[\boldsymbol{\theta} \sim \text{Dirichlet}(\boldsymbol{\alpha}_i).\]
   Here, \(\boldsymbol{\theta}\) represents a probability distribution over available modalities.

3. \textbf{Sampling Tokens from the Modalities:}  
   Finally, given \(\boldsymbol{\theta} = (\theta_1, \theta_2, \ldots, \theta_n)\), the number of tokens for each modality \(i\) is \(T_i = 2048 \times \theta_i\). Within this cap, tokens from each modality are sampled randomly. 2048 is the maximum number of input and target tokens.

\subsection{EgoM2P Pretraining Details}
\label{sup:training}

For the model architecture, we adapt the T5-Base model. It has 12 Transformer blocks in both the encoder and decoder. The latent dimension is 768. The model uses 12 attention heads in its multi-head attention mechanism. For each modality embedding layer, it maps each token in the codebook to a 768-dimensional space.
% The concentration parameters for 
In Tab. \ref{tab:main_model}, we detail the training hyperparameters. The model is trained with distributed data parallel. 

The total number of parameters of \methodname~is approximately 400 million. 
% During training, we first randomly sample a dataset with probability proportional to its size; then, similar to 4M, we use Dirichlet distribution to sample visible input tokens and target tokens from available multimodalities. 
During training, the maximum number of sampled tokens for both input and target is 2048. The total number of training tokens is 400 billion, randomly sampled and masked from our database of 4 billion tokens. 

The model training takes 16 hours using 256 NVIDIA GH200 superchips.  All networks, including tokenizers, are trained from scratch without initializing parameters from existing large models.

\begin{table}[t]
\centering
\begin{tabular}{l c}
\toprule
\textbf{Configuration} & \textbf{EgoM2P} \\
\midrule
Training tokens & 400B \\
Warmup tokens & 10B \\
Optimizer & AdamW~\cite{loshchilov2018decoupled} \\
Opt. momentum & $\beta_1, \beta_2 = 0.9, 0.99$ \\
Base learning rate & $1\text{e-}4$ \\
Total batch size & 1024 \\
Weight decay & 0.05 \\
Max gradient norm & 1 \\
Learning rate sched. & Cosine decay \\
% Feedforward activation & SwiGLU~\cite{ref100} \\
\midrule
Max input token number & 2048 \\
Max target token number & 2048 \\
% Input and target $\alpha$  & 0.5, 0.5 \\
% Masking strategy & RGB $\rightarrow$ All \& All $\rightarrow$ All \\
Data type & bfloat16 \\
\bottomrule
\end{tabular}
\caption{\textbf{Pretraining settings.}}
\label{tab:main_model}
\end{table}

\begin{algorithm}[t]
\caption{Handling Unaligned Multimodalities}
\begin{algorithmic}
    \State \textbf{Input:} Data iterators $\{D_i\}_{i=1}^N$, dataset sampling probabilities $\mathbf{p} = \{p_i\}_{i=1}^N$, training modalities $\{m_j\}_{j=1}^M$
    \State $i \sim \text{Categorical}(\mathbf{p})$ \hfill \Comment{Sample dataset index $i$}
    % \State $\mathbf{x} \gets \text{iter}(D_i)$ \Comment{Contains \# of modalities $\leq M$}
    \State \textit{/* Sample input and target tokens for i (Sec. A.2) */}
    \State $\boldsymbol{\alpha} \sim [\boldsymbol{\alpha}_1, \boldsymbol{\alpha}_2, \boldsymbol{\alpha}_3, \boldsymbol{\alpha}_4]$   \Comment{Sample Dir. concentr. param}
    \State $\boldsymbol{\alpha} \gets \text{Adjust}(\boldsymbol{\alpha}, \text{missing modalities})$ \hfill \Comment{Exclude params for missing modalities}
    \State Sample input token counts $ic$ for each modality \Comment{Ensure modality token limits per sample}
    \State Sample target token counts $tc$ \Comment{Modality token limits per sample are adjusted by subtracting $ic$}
    \State $\mathbf{im} \gets \mathbf{1}, \mathbf{tm} \gets \mathbf{1}$ \Comment{Init input and target mask}
    \State Random sample $ic$ tokens, set input mask $\mathbf{im}$ to 0
    \State Random sample $tc$ tokens (non-overlapping with $ic$), set target mask $\mathbf{tm}$ to 0
    \State $\mathbf{x} \gets \{\text{data}, \mathbf{im}, \mathbf{tm}\}$ \Comment{For each existing modality}
    \State $\mathbf{x'} \gets \{\}$ 
    \For{$j = 1$ to $M$} \hfill \Comment{Iterate over $M$ modalities}
        \State $\mathbf{x}'[j][\text{data}] \gets \mathbf{0}$ \hfill \Comment{Tensor initialized to $0$}
        \State $\mathbf{x}'[j][\mathbf{im}] \gets \mathbf{1}$ \hfill \Comment{Denotes not used in input tokens}
        \State $\mathbf{x}'[j][\mathbf{tm}] \gets \mathbf{1}$ \hfill \Comment{Denotes not used in target tokens}
    \EndFor
    \State $\mathbf{x'}$.update($\mathbf{x}$) \Comment{Replace placeholders with real data with random input/target masks sampled by the Dirichlets}
    \State Select 2048 input and target tokens with priority given to $\mathbf{im}=0$ \text{and} $\mathbf{tm}=0$
    \State No \textit{Enc.} self-attention and cross-attention when $\mathbf{im} = 1$
    \State \textit{Dec.} self-attention processes visible same-modal tokens
\end{algorithmic}
\label{alg:missing_modality}
\end{algorithm}

\subsection{Missing Modality Handling Details}
Unlike 4M, which relies on an aligned multimodal dataset containing all modalities, our approach uses missing modality masking. This allows us to scale training across multiple real-world multimodal egocentric datasets, even when the modalities are unaligned. See Alg. \ref{alg:missing_modality} for the pseudo-code.

\subsection{Inference Details}

As an order-agnostic autoregressive model, \methodname~supports parallel decoding in each decoding step. All target tokens can be decoded at the same time, however, increasing the decoding step $s$ to 3 to 6 generally produces higher-quality predictions. In each decoding step, previously decoded target tokens are conditioned to ensure prediction consistency. See Alg. \ref{alg:inference} for the pseudo-code.

\begin{algorithm}[t]
\caption{\methodname~Inference}
\begin{algorithmic}
    \State \textbf{Input:} input tokens $\mathbf{I}$, target modality $t$, decoding steps $s$, target token number $n$, guidance scale $w$
    % \State $\mathbf{y} \gets [y_1, y_2, 0]$ \Comment{Project to the xy plane}
    \State $\mathbf{T} \gets \mathbf{0}$ \Comment{Init target tokens prediction}
    \State $\mathbf{im} \gets \mathbf{1}$ \Comment{Init input mask for $t$. 1: not used as input}
    \State $\mathbf{tm} \gets \mathbf{0}$ \Comment{Init target mask for $t$. 0: need to predict}
    \For{$j = 1$ to $s$}
        \State $n_j \gets n / s$ \Comment{Num of tokens to decode in this step}
        \State \textit{/* Pass 1: conditional distribution prediction */}
        \State context $\gets$ \textit{Enc}($\{\mathbf{I}, \mathbf{T}[\mathbf{1} - \mathbf{im}]\}$)
        \State $\mathbf{s}_j \gets \text{Sample}(n_j, \{i \mid \mathbf{tm}[i] = 0\})$ \Comment{Randomly sample $n_j$ indices from unpredicted tokens}
        \State $\mathbf{p}_{\text{cond}}[\mathbf{s}_j] \gets \textit{Dec}(\mathbf{T}[\mathbf{s}_j], \text{context})$
        \State \textit{/* Pass 2: unconditional distribution prediction */}
        \State context' $\gets$ \textit{Enc}($\{\mathbf{T}[\mathbf{1} - \mathbf{im}]\}$) \Comment{Mask all input tokens}
        \State $\mathbf{p}_{\text{uncond}}[\mathbf{s}_j] \gets \textit{Dec}(\mathbf{T}[\mathbf{s}_j], \text{context'})$
        \State $\mathbf{p}[\mathbf{s}_j] \gets \mathbf{p}_{\text{uncond}} + (\mathbf{p}_{\text{cond}} - \mathbf{p}_{\text{uncond}}) * w$ \Comment{CFG}
        \State $\mathbf{T}[\mathbf{s}_j] \gets$ Nucleus Sampling $\sim \mathbf{p}[\mathbf{s}_j]$
        \State $\mathbf{im}[\mathbf{s}_j] \gets \mathbf{0}$
        \State $\mathbf{tm}[\mathbf{s}_j] \gets \mathbf{1}$
    \EndFor
    \State \textbf{Return:} $\mathbf{T}$ \Comment{Target tokens prediction} 
\end{algorithmic}
\label{alg:inference}
\end{algorithm}

\section{Ablation Studies}
\label{sup:ablation}

\subsection{Number of Visible Tokens}
First, we do an ablation study on the maximum number of visible input and target tokens. 4M \cite{4m, 4m21} set this to 256 and argue that ``the challenge of the multimodal masked modeling task is mainly determined by how many visible input tokens are used; having fewer tokens makes the task more difficult. This is because the modalities provide a lot of spatial information about each other, so it's important to reduce the number of visible tokens to keep the task challenging enough.'' However, in our task, each video sample has more than 5000 tokens. We find that increasing the maximum input and target token numbers helps the multimodal masked modeling on videos. We follow 4M to report the validation set loss as metrics to show how well the pretraining is. Refer to Tab. \ref{tab:input_masking_budget}. In order to balance the maximum number of tokens and training efficiency, we choose the maximum number of input/target tokens as 2048.

\subsection{Dataset Sampling Weights}
Secondly, in our experiment, we observed that the sampling weights assigned to different datasets and modalities significantly impact both training stability and model performance. This is particularly important because our datasets are highly imbalanced; for instance, EgoExo4D \cite{Grauman2023EgoExo4DUS} contains 160 times more samples than H2O \cite{Kwon_2021_ICCV}. Training a large model on such skewed datasets can result in two major issues: Smaller datasets may be overlooked, or smaller datasets may suffer from overfitting.

To address these challenges, we found that sampling datasets with probabilities proportional to their sizes helps achieve better balance. To further explore this, we conducted three ablation studies:
\begin{enumerate}
    \item Sampling datasets based on probabilities proportional to their sizes.
    \item Sampling datasets using a uniform distribution.
    \item Sampling datasets with probabilities proportional to the logarithm of their sizes.
\end{enumerate}

We observe that across all datasets in our database, the first sampling method, which selects datasets based on probabilities proportional to their sizes, consistently results in the lowest validation loss. For large-scale datasets such as EgoExo4D, the third method, which samples datasets with probabilities proportional to the logarithm of their sizes, achieves a lower validation loss compared to the second method, which employs a uniform distribution. However, both the second and third methods tend to overfit during the early stages of training on smaller-scale datasets. 

\begin{table}[t]
\centering
\begin{tabular}{c|c}
\toprule
\textbf{Input/Target Tokens} & \textbf{Avg. Loss} \\
\midrule
1024 & 5.80\\
\textbf{2048} & \textbf{4.93} \\
\bottomrule
\end{tabular}
\caption{\textbf{Maximum visible token ablation.}}
\label{tab:input_masking_budget}
\end{table}

\subsection{EgoGen \cite{li2024egogen} Contributions}
Collecting large-scale egocentric datasets with accurate 3D ground-truth annotations is expensive, time-consuming, and non-scalable. Compared to the internet-scale third-person view video data, EgoGen \cite{li2024egogen} offers a practical solution to scale up the training data of egocentric foundation models. We perform ablation studies on the contributions of egocentric synthetic data from EgoGen \cite{li2024egogen}: We remove EgoGen from the training set, fix other settings, and test the model on the same test set. In Tab. \ref{tab:ablation-egogen-cam} and \ref{tab:ablation-egogen-depth}, we show that cheap and high-quality egocentric synthetic data can boost performance on egocentric camera tracking and egocentric video depth estimation, complementing expensive real data.

\begin{table}[t]
    \centering
    \setlength{\tabcolsep}{3.5pt} % Reduce column spacing
    \renewcommand{\arraystretch}{1.} % Adjust row spacing
    \scriptsize
    \begin{tabular}{l|ccc|ccc}
        \toprule
        \multirow{2}{*}{Method} & \multicolumn{3}{c|}{EgoExo4D~\cite{Grauman2023EgoExo4DUS}} & \multicolumn{3}{c}{ADT~\cite{pan2023ariadigitaltwinnew} (\textit{unseen})}  \\
        & ATE$\downarrow$ & RTE$\downarrow$ & RRE$\downarrow$ & ATE$\downarrow$ & RTE$\downarrow$ & RRE$\downarrow$ \\
        \midrule
        \multirow{1}{*}{\methodname~w/o EgoGen} & 0.028 & 0.005 & 0.561 & 0.053 & 0.009 & 0.593 \\ 
        \midrule
        \multirow{1}{*}{\methodname} & \textbf{0.017} & \textbf{0.004} & \textbf{0.429} & \textbf{0.032} & \textbf{0.006} & \textbf{0.490} \\
        \bottomrule
    \end{tabular}
    \caption{\textbf{Ablation of EgoGen \cite{li2024egogen} on camera tracking.}}
    \label{tab:ablation-egogen-cam}
\end{table}

\begin{table}[t]
    \centering
    \setlength{\tabcolsep}{3pt} % Reduce column spacing
    \renewcommand{\arraystretch}{1.} % Adjust row spacing
    \scriptsize
    \begin{tabular}{l|cc|cc}
        \toprule
        \multirow{2}{*}{Method}
        & \multicolumn{2}{c|}{H2O~\cite{Kwon_2021_ICCV}} & \multicolumn{2}{c}{HOI4D~\cite{Liu_2022_CVPR} (\textit{unseen})} \\
        & Abs Rel $\downarrow$ & $\delta_{1.25}\uparrow$ & Abs Rel $\downarrow$ & $\delta_{1.25}\uparrow$ \\
        \midrule
        \multirow{1}{*}{\methodname~w/o EgoGen} & 0.062 & 94.9 & 0.067 & 97.1 \\
        \midrule
        \multirow{1}{*}{\methodname} & \multirow{1}{*}{\textbf{0.055}} & \multirow{1}{*}{\textbf{96.0}} & \textbf{0.061} & \textbf{98.0} \\
        \bottomrule
    \end{tabular}
    \caption{\textbf{Ablation of EgoGen \cite{li2024egogen} on depth estimation.}}
    \label{tab:ablation-egogen-depth}
\end{table}

\section{Tokenization Error Analysis}
\methodname~predicts tokens in a quantized form, which leads to the propagation of quantization errors throughout the pipeline. To examine the impact of tokenizers, we present the reconstruction errors in Table \ref{tab:quantizationerror}. We compare the quantization error with the \methodname~prediction error in the egocentric camera tracking and gaze estimation tasks. Quantization error is not the bottleneck in our model for Euclidean data. The relatively high rotation error (RRE) highlights challenges of VQ for 4D egocentric data in non-Euclidean space.

\begin{table}[t]
    \centering
    \setlength{\tabcolsep}{3.5pt} % Reduce column spacing
    \renewcommand{\arraystretch}{1} % Adjust row spacing
    \scriptsize
    \begin{tabular}{l|ccc|c}
        \toprule
         & \multicolumn{3}{c|}{EgoExo4D (cam tracking)}  &  \multicolumn{1}{c}{EgoExo4D (gaze pred.)}\\
        &ATE$\downarrow$&RTE$\downarrow$&RRE$\downarrow$& MSE$\downarrow$\\
        \midrule 
        \methodname & 0.017 & 0.004 & 0.429  & 0.0162\\
        Quantization error & 0.005 & 0.001 & 0.272 & 0.0000188\\
        \bottomrule
    \end{tabular}
    \caption{\textbf{Quantization error analysis.}} 
    \label{tab:quantizationerror}
\end{table}

\section{Post-Training Details}
With the evolving quantity of egocentric datasets, \methodname~can be easily adapted to unseen datasets by post-training. In the main paper (Sec. 4.5), we leverage post-training on the training sets of unseen datasets, including ADT \cite{pan2023ariadigitaltwinnew}, HOI4D \cite{Liu_2022_CVPR}, and ASE \cite{ase}, and demonstrate that the post-trained \methodname~outperforms SOTA specialist baselines. The ADT training set consists of 10,885 paired RGB and camera trajectory samples, while the HOI4D set includes 11,740 paired RGB and depth video samples. Additionally, the ASE set contains 26,000 paired RGB and depth video samples. All samples are randomly selected from the original dataset, ensuring no overlap with the test set. All modalities are standardized as described in Sec.~3.1. We initialize the post-training with the pretrained \methodname, and continue to train it for 50B tokens. The number of warmup tokens is 5B, and other settings are the same as Tab. \ref{tab:main_model}. We observe that post-training on ADT \cite{pan2023ariadigitaltwinnew} tends to overfit more quickly compared to the other two datasets. Consequently, we select the best model for each task based on its respective validation loss during post-training.

\section{Egocentric 4D Reconstruction}
% predicts depth video and camera trajectory efficiently from monocular egocentric videos. Given ground-truth camera intrinsics and an egocentric video, \methodname~performs 4D reconstruction $\sim$70 times faster than the SOTA baseline MegaSAM \cite{li2024_MegaSaM}. Unlike MegaSAM, which relies on SOTA monocular depth estimators and expensive geometry optimization, EgoM2P efficiently reconstructs dynamic egocentric scenes.
Given ground-truth camera intrinsics and an egocentric video, we compare \methodname~with the SOTA baseline MegaSaM \cite{li2025megasam} for 4D reconstruction. Unlike MegaSaM, which relies on SOTA monocular depth estimators and expensive geometry optimization, \methodname~efficiently reconstructs dynamic egocentric scenes. For a 2-second video at 8 FPS, \methodname~completes the reconstruction in less than 1 second, whereas MegaSaM requires 71 seconds. We provide a qualitative comparison in Fig. \ref{fig:4d}.

\begin{figure}[t]
    \centering
    \begin{minipage}[b]{0.23\textwidth}
        \includegraphics[width=\textwidth]{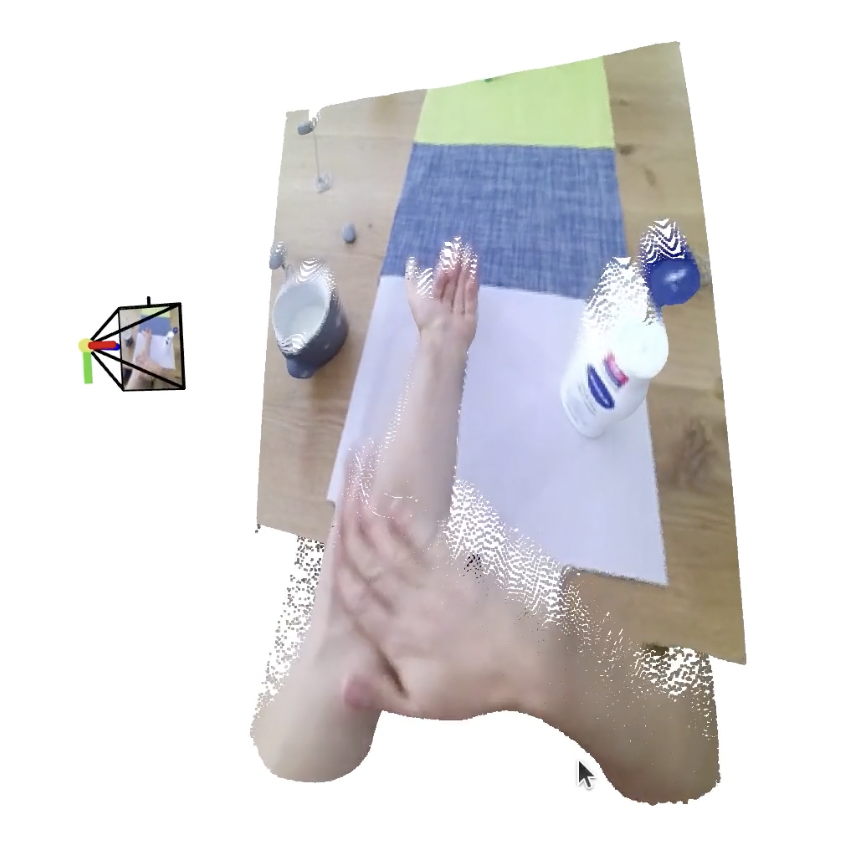}
        \caption*{MegaSaM (71 s)}
        \label{fig:figure1}
    \end{minipage}
    \hfill
    \begin{minipage}[b]{0.23\textwidth}
        \includegraphics[width=\textwidth]{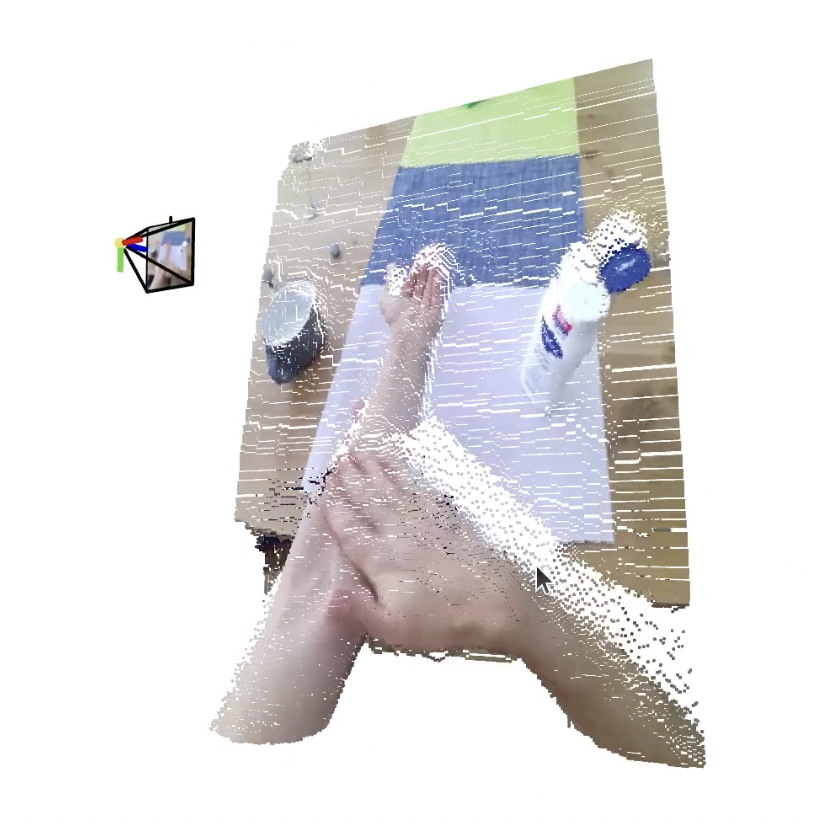}
        \caption*{\methodname~($<$1 s)}
        \label{fig:figure2}
    \end{minipage}
    \caption{Dynamic 4D Reconstruction from Monocular Egocentric Videos.}
    \label{fig:4d} % Label for the entire figure environment
\end{figure}

\end{document}